%% file: sample-acmtog-SIGGRAPH-submission.tex
\documentclass[acmtog]{acmart}
\acmSubmissionID{1233}

\usepackage{booktabs} 

\usepackage{wrapfig}
\usepackage{overpic}
\usepackage{graphicx}

\usepackage[ruled]{algorithm2e} 

\SetAlFnt{\small}
\SetAlCapFnt{\small}
\SetAlCapNameFnt{\small}
\SetAlCapHSkip{0pt}

\let\maketitlesup\maketitle
\usepackage{xpatch}
\xpatchcmd{\maketitlesup}{\@mkteasers}{}{}{}
\xpatchcmd{\maketitlesup}{\@mkabstract}{}{}{}
\xpatchcmd{\maketitlesup}{\@mkbibcitation}{}{}{}
\xpatchcmd{\maketitlesup}{\@mklinecount}{}{}{}
\xpatchcmd{\maketitlesup}{\@keywords}{}{}{}
\xpatchcmd{\maketitlesup}{\@mkauthors}{}{}{}
\xpatchcmd{\maketitlesup}{\@authorsaddresses}{}{}{}
\xpatchcmd{\maketitlesup}{\@authornotes}{}{}{}

\acmJournal{TOG}

\copyrightyear{2024}
\acmYear{2024}
\setcopyright{rightsretained}
\acmConference[SA Conference Papers '24]{SIGGRAPH Asia 2024 Conference Papers}{December 3--6, 2024}{Tokyo, Japan}
\acmBooktitle{SIGGRAPH Asia 2024 Conference Papers (SA Conference Papers '24), December 3--6, 2024, Tokyo, Japan}\acmDOI{10.1145/3680528.3687700}
\acmISBN{979-8-4007-1131-2/24/12}

\begin{document}
\title{NASM: Neural Anisotropic Surface Meshing}

\author{Hongbo Li}
\affiliation{  \institution{Wayne State University}
\country{USA}}\email{hm9026@wayne.edu}

\author{Haikuan Zhu}
\affiliation{  \institution{Wayne State University}
\country{USA}}\email{hkzhu@wayne.edu}

\author{Sikai Zhong}
\affiliation{  \institution{Wayne State University}
\country{USA}}\email{sikai.zhong@wayne.edu}

\author{Ningna Wang}
\affiliation{  \institution{The University of Texas at Dallas}
\country{USA}}\email{ningna.wang@utdallas.edu}

\author{Cheng Lin}
\affiliation{  \institution{The University of Hong Kong}
\country{China}}\email{chlin@connect.hku.hk}

\author{Xiaohu Guo}
\affiliation{  \institution{The University of Texas at Dallas}
\country{USA}}\email{xguo@utdallas.edu}

\author{Shiqing Xin}
\affiliation{  \institution{Shandong University}
\country{China}}\email{xinshiqing@sdu.edu.cn}

\author{Wenping Wang}
\affiliation{  \institution{Texas A\&M University}
\country{USA}}\email{wenping@tamu.edu}

\author{Jing Hua}
\affiliation{  \institution{Wayne State University}
\country{USA}}\email{jinghua@wayne.edu}

\author{Zichun Zhong}
\authornote{Corresponding author.}
\affiliation{  \institution{Wayne State University}
\country{USA}}\email{zichunzhong@wayne.edu}

\renewcommand{\shortauthors}{Li, Zhu, Zhong, Wang, Lin, Guo, Xin, Wang, Hua, and Zhong}

\begin{abstract}
This paper introduces a new learning-based method, NASM, for anisotropic surface meshing. Our key idea is to propose a graph neural network to embed an input mesh into a high-dimensional (high-d) Euclidean embedding space to preserve curvature-based anisotropic metric by using a dot product loss between high-d edge vectors. This can dramatically reduce the computational time and increase the scalability. Then, we propose a novel feature-sensitive remeshing on the generated high-d embedding to automatically capture sharp geometric features. We define a high-d normal metric, and then derive an automatic differentiation on a high-d centroidal Voronoi tessellation (CVT) optimization with the normal metric to simultaneously preserve geometric features and curvature anisotropy that exhibit in the original 3D shapes. To our knowledge, this is the first time that a deep learning framework and a large dataset are proposed to construct a high-d Euclidean embedding space for 3D anisotropic surface meshing. Experimental results are evaluated and compared with the state-of-the-art in anisotropic surface meshing on a large number of surface models from Thingi10K dataset as well as tested on extensive unseen 3D shapes from Multi-Garment Network dataset and FAUST human dataset.
\end{abstract} 

\begin{teaserfigure}
\centering
   \includegraphics[width=0.9\textwidth]{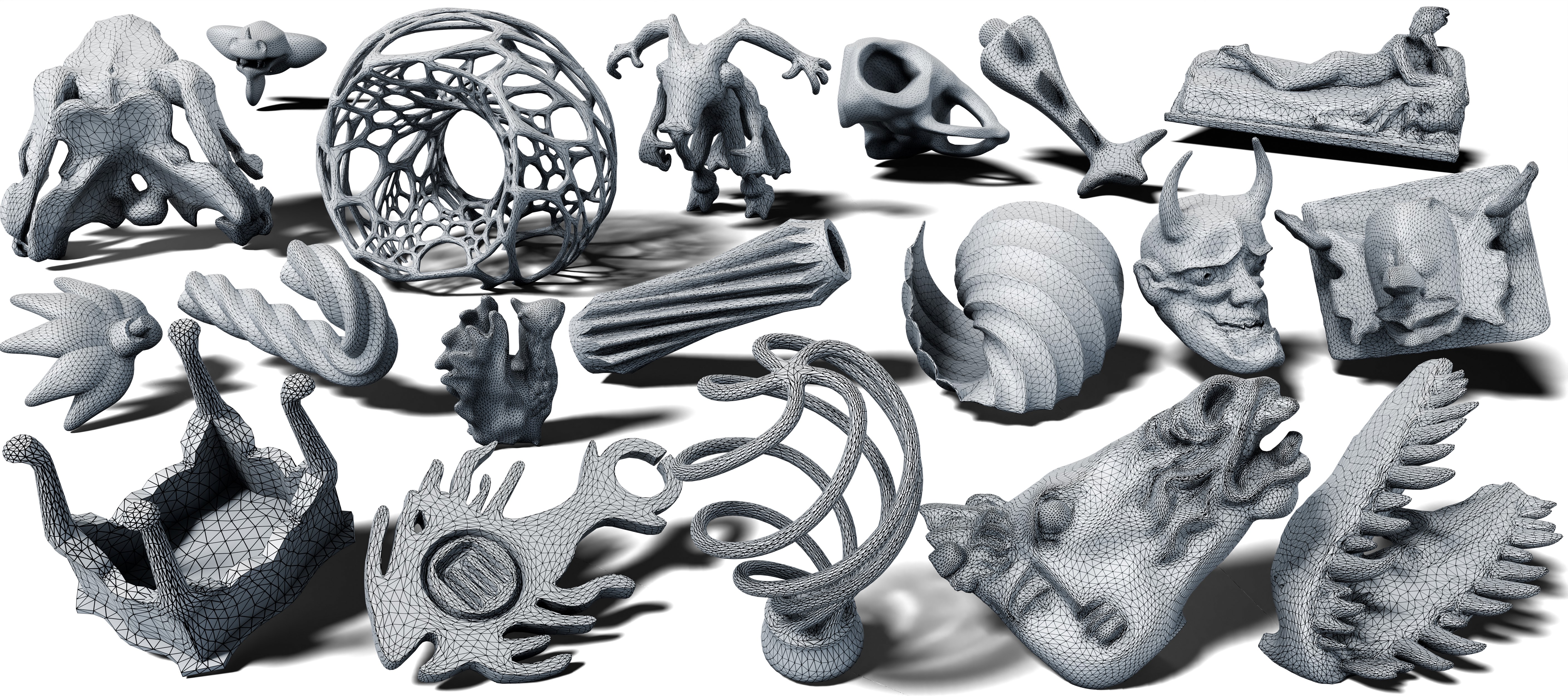} 
   \caption{A gallery of anisotropic surface meshes generated by our NASM method. These results are selected from our testing models in Thingi10K dataset, including complicated organic surfaces, and surfaces with sharp and weak features as well as varying anisotropic metrics.}
\label{fig:teaser}
\end{teaserfigure}

%
%


%
%


\maketitle

\input{samplebody-journals}

\end{document}

%% file: samplebody-journals.tex
\input{section/1_introduction}
\input{section/2_related_work}
\input{section/3_neural_highd_embedding}
\input{section/4_normal_metrci_cvt}
\input{section/5_experiment}

\section{Conclusion}
In this article, we develop a novel scalable anisotropic surface mesh generation method. To our knowledge, this is the first deep learning-based method capable of generating a large number of high-quality and high-fidelity anisotropic surface meshes. There are several advantages of this method compared with traditional methods: (1) there is no curvature metric needed as input for our mesh generation. It can relieve the issues coming from the curvature estimation; (2) the high-d embedding computational time has been dramatically reduced to real time with the help of the designed learning-based method; (3) the developed high-d normal metric CVT formulation can generate feature-sensitive anisotropic meshes, which well capture both sharp and weak features; (4) this method is robust to generate a large number of anisotropic surface meshes from complicated geometric shapes.\vspace{-1mm}

\section{Limitations and Future Work} 
For some CAD-like models with very sparse input vertices and flat planes only, NASM may fail, since the high-quality embeddings are very challenging to be computed / predicted in such cases, which have been shown in Section F.3 of Supplemental Document. Another limitation is about the generalization to other anisotropic metrics fields. This requires the preparation of such training datasets for those particular applications. In the future, we will extend the current framework to deal with large-scale 3D scene mesh generation. It is also promising to explore how to effectively integrate the high-d normal metric CVT computation into the neural network computing framework. Since curvature-induced anisotropic meshes are essential for enhancing accuracy, efficiency, and stability in simulations involving complex geometries across various fields, we will apply our method in fluid dynamics, computer animation, and medical simulations, etc.\vspace{-2mm}

\begin{acks}
The authors would like to thank the anonymous reviewers for their valuable comments and suggestions. Hongbo Li, Haikuan Zhu, Sikai Zhong, Jing Hua, and Zichun Zhong were partially supported by National Science Foundation (OAC-1845962, OAC-1910469, and OAC-2311245). Ningna Wang and Xiaohu Guo were partially supported by National Science Foundation (OAC-2007661). Shiqing Xin was partially supported by National Key R\&D Program of China (2021YFB1715900). 

\end{acks}

\bibliographystyle{ACM-Reference-Format}
\bibliography{sample-bibliography}

\clearpage
\input{section/appendix}

%% file: section/1_introduction.tex
\section{Introduction}
In geometric modeling, physical simulation, and mechanical engineering fields, anisotropic meshes are crucial for performing better shape approximations~\cite{Simpson:1994} and achieving higher accuracy in numerical simulations~\cite{Alauzet:JCP2010,Narain:TOG2012}. Anisotropic surface meshes are triangulations with elements elongated along prescribed directions and stretchings. One of the fundamental geometric merits is that, with a given number of mesh elements (i.e., vertices or triangles), the $L_2$ optimal approximation to a smooth surface is achieved when the anisotropy of triangles conforms to the eigenvalues and eigenvectors of the curvature tensors~\cite{Simpson:1994,Heckbert:1999}. This improves the simulation’s fidelity, stability, and efficiency. They are vital for high-fidelity structural analysis, such as in the study of turbine blades, aircraft wings, or biomedical implants. Due to the difficulty in anisotropic meshing, recently some researchers~\cite{Particle2013,Embedding2018} proposed a computational method to map the complicated anisotropic 3D space onto a higher dimensional Euclidean space, which can make the anisotropic mesh generation computation simpler. This research line is inspired by Nash Embedding Theory~\cite{Nash:1954,Kuiper:1955}. However, the major bottleneck of the high-dimensional (high-d) embedding method is time-consuming in the computation. Some other high-d embedding-based methods can only use specific embeddings, such as normal-based embedding, which cannot consider arbitrary input metrics~\cite{levy2013variational,DASSI2014,DASSI2015}. 

Besides that, most existing anisotropic meshing methods~\cite{Bossen1996,Alliez2003,Du2005,valette2008,Particle2013,fu2014anisotropic,boissonnat2015a,Embedding2018} need to have a given metric as input to control the element stretching ratio and orientation, which is tedious and unrobust. Furthermore, in order to handle geometric sharp or weak features, users need to identify these feature edges and corners in the input reference mesh at first, and then constrain the mesh vertex position and connectivity along the feature edges or fix the feature corners during optimization. \cite{levy2010LpCVT} and~\cite{xu2024cwf} extend the CVT objective function by a metric term, which can automatically attract the site point onto the feature lines in 3D space so as to naturally recover the features. However, it can only handle with isotropic meshing. To our knowledge, there is no method which can generate anisotropic mesh to preserve both metric field as well as sharp / weak features. Moreover, all the existing methods for anisotropic meshing are model-based approaches, which are not scalable to generate a large number of anisotropic meshes from a variety of shape geometries and typologies. 

In this work, we address the above-mentioned challenges in anisotropic mesh generation. It includes main twofold: (1) how to develop a learning-based method to efficiently and robustly compute a high-d embedding without providing a pre-computed metric field; (2) how to generate high-fidelity and high-quality anisotropic surface meshes with automatical feature preserving. The \textit{main contributions} are as follows:
\begin{itemize}
  \item Develop a scalable computational paradigm to generate high-quality anisotropic surface meshes from arbitrary input meshes only (no curvature metric is needed);
 \item Design an efficient GNN-based method with high-d dot product loss to embed an input mesh into a high-d Euclidean embedding space to preserve curvature-based anisotropic metric (a speedup of about $1,500\times$ times);
 \item Define a high-d normal metric CVT optimization with an automatic differentiation to compute the feature-sensitive anisotropic surface meshes (without any user's input on tagging features);
 \item Construct a large dataset for anisotropic surface meshing and processing (more than 800+ mesh models from Thingi10K, Multi-Garment Network, and FAUST datasets). 
\end{itemize}

The overview pipeline of the proposed neural anisotropic surface meshing (NASM) approach is shown in Fig.~\ref{fig:pipeline}.

%% file: section/2_related_work.tex
\section{Related Works}
In this section, we review the related literature on anisotropic triangle meshing approaches and neural geometric learning on meshes.  
\begin{figure*}[t!]
    \centering
    \includegraphics[width=0.88\textwidth]{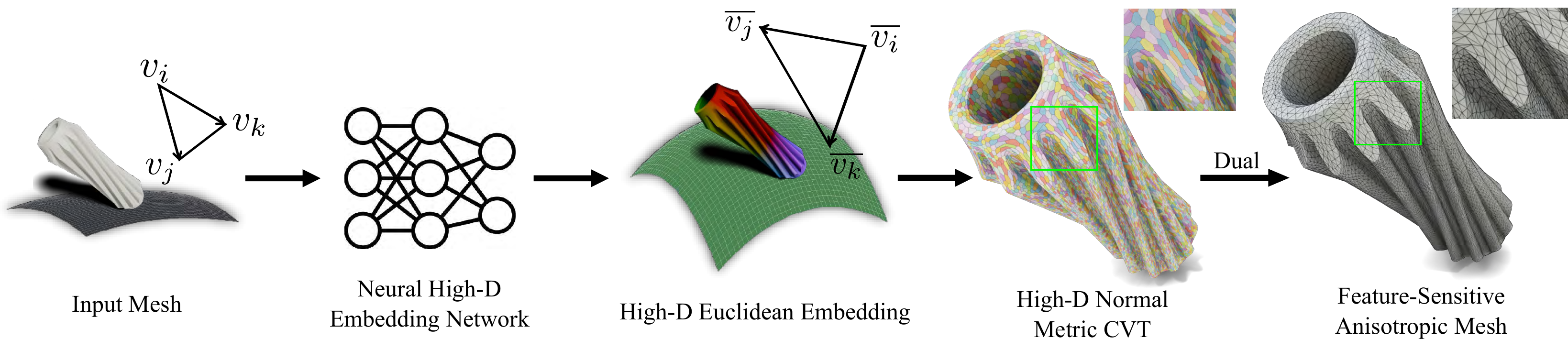}\vspace{-2mm}
    \caption{The overview pipeline of the proposed neural anisotropic surface meshing (NASM) approach. Our method includes two main components: neural high-d Euclidean embedding and high-d normal metric CVT for feature-sensitive anisotropic meshing. The training data for neural high-d Euclidean embedding is generated based on $\text{SIFHDE}^2$~\cite{Embedding2018}. More details are given in Section A of Supplemental Document.}\vspace{-2mm}
    \label{fig:pipeline}
\end{figure*}

\subsection{Anisotropic Triangle Meshing}
In terms of meshing a surface, anisotropic is related to optimal approximation of a discretization of the surface function with a given number of element count~\cite{Alliez2003}. Anisotropic triangular meshing has been extensively studied over decades, both on flat, 2D regions~\cite{Bossen1996}, and the Euclidean 3D surface~\cite{shimada1997}. \cite{shapiro1996} introduces the Adaptive Smoothed Particle Hydrodynamics (ASPH) which uses inter-particle Gaussian kernels with an anisotropic metric tensor. \cite{Particle2013} further extends the formulation of the energy between particles to a higher embedding space which leads to enhanced flexibility and accuracy when confronted with either mild or significant variations in the metric.


\subsubsection{Anisotropic Centroidal Voronoi Tessellation}
\cite{Du2005} further generalizes the concept of CVT to the anisotropic centroidal Voronoi tessellation (ACVT) by integrating the given Riemannian metric into the calculation. However, the Riemannian metric needs to be constructed in each Lloyd~\cite{Lloyd1982} iteration, which is not efficient. \cite{valette2008} provides a discrete approximation of ACVT to accelerate the computation speed with the cost of degraded mesh quality. \cite{zhong2014anisotropic} computes the CVT on an 2D parametric domain where the metric surface can be conformally mapping to. \cite{levy2013variational} leverages the embedding theory~\cite{Nash:1954} and formulates a 6D space where the surface mesh can be embedded in, followed with the CVT isotropic remeshing within this extended dimensionality. The final results reveal the distinctive anisotropic characteristics, albeit not distributed across the entire surface. \cite{Embedding2018} introduces a variational approach to compute the higher dimensional space through aligning the Jacobian of transformation and the gradient of deformation between the 3D space and the high-d space where the surface is embedded. This approach considers the metric space defined over the surface which leads the result with anisotropic properties regarding to the given metric. However, the computation is involved in solving a linear system w.r.t. the input mesh resolution, which demands a significant investment of time.

\subsubsection{Anisotropic Delaunay Triangulation}
Many works have extended the Delaunay triangulation to the anisotropic case, both through the refinement-based approaches~\cite{frey1999surface,dobrzynski2008} and the variational approaches~\cite{chen2004optimal,chen2007optimal}. \cite{boissonnat2015a,boissonnat2015b} leverages a Delaunay refinement scheme that progressively inserts vertices to reach the final anisotropic meshes. \cite{rouxel2016discretized} implements an algorithm for the computation of the discrete approximations to Riemannian Voronoi diagrams on 2-manifolds employing numerical methods for calculating the geodesic distances. 
\cite{fu2014anisotropic} presents a Locally Convex Triangulation (LCT) method to integrate anisotropic into optimal Delaunay triangulation, which constructs convex functions that locally match the predefined Riemaninain metric. \cite{budninskiy2016optimal} introduces a variational method to lift the points onto a convex function, and minimize the error between the lifted function and the constructed mesh. The utilization of the convex function instead of commonly used paraboloid integrates the anisotropy in final result. However, their limitation is only applicable to a small class of anisotropies that can be represented by convex functions.

\subsubsection{Mesh Processing on High Dimensional Space}
As every smooth Riemannian manifold can be isometrically embedded into some Euclidean space~\cite{Nash:1954}, several mesh processing tasks related to Riemannian manifold have been studied in 3D or higher dimensions. \cite{panozzo2014frame} computes a 3D embedding via surface deformation to obtain anisotropic meshing results. \cite{DASSI2014} follows the 6D configurations the same as in~\cite{levy2013variational} and proposed local mesh operations to better preserve important geometric features on CAD models. \cite{DASSI2015} configures the extended dimensions with a smooth function or the solution of a partial differential equation, whereas they consider only the case in the 2D space. \cite{Embedding2018} recently proposes to directly calculate the high dimension counterpart and use the coordinates of original dimension as the first three dimensions to achieve a self-intersection free high-d embedding mesh.

\subsection{Neural Geometric Learning on Meshes}
\subsubsection{Non-Graph-Based Neural Network}
Recent advances in deep learning facilitate the use of learnable components in 3D modeling applications. One way of applying learnable methods is to treat 3D geometry models using regular data structures, whereupon the established learning paradigms in the 2D image domain can be seamlessly extended, providing a straightforward mean of processing and analyzing. For instance, geometry models can be described with 3D grids of values and 3D Convolutional Neural Network (CNN) can be applied~\cite{wang2017cnn,wang2018adaptive,mescheder2019occupancy,Chen_2021}. The other way is to directly handle the irregular inherence of the geometry structure. An earlier work~\cite{ivrissimtzis2004neural} proposes a learning algorithm for surface reconstruction by simulating an incrementally expanding neural network which learns a point cloud through a learning process. Recently, PointNet~\cite{qi2017pointnet} and PointNet++~\cite{Qi_pointnet++} stand as pioneering methods in this direction. Nevertheless, lack of explicit connectivity information acquires intensive computational demands and limited representational capacity for the point structure. Surface mesh, as a conventional widely adopted structure, has been used for neural geometric learning and manifest favorable prospects. Recent surveys~\cite{Bronstein_2017,xiao2020survey} can be referred.

\subsubsection{Graph-Based Neural Network}
A common representation for irregular data is the graph structure. As mesh data being a special type of graph structure, several works have advanced to leverage graph-based neural networks (GNNs)~\cite{kipf2017semi,hamilton2018inductive} and their spectral variants~\cite{Bruna2013,Defferrard2016,Kostrikov2018} on classic discrete mesh problem. \cite{Sharp2022DiffusionNet,Smirnov2021} extend the Laplacian operator to apply learning on the graph. \cite{Hu2021} introduces a uniform downsampling scheme into the learning pipeline. \cite{pfaff2020} leverages the GNN on simulation tasks. \cite{hanocka2019} constructs an edge-based graph and defines convolution on it. \cite{Wang2020} generates 3D shape from 2D by updating a coarse mesh through GNN. \cite{Potamias2022} leverages GNN to predict the connectivity in mesh simplification. The closest application to our work is~\cite{Pang2023}, which exploits the neural network to map the meshes to high-d embedding and calculate the geodesic on the surface. However, their method is not designed for learning Riemannian metric and their loss function is not effective on anisotropic mesh generation. We conduct experiments to show that the loss functions as they proposed, e.g., L1 or L2 loss, present inferior quality in our meshing results.

%% file: section/3_neural_highd_embedding.tex
\section{Neural High-D Euclidean Embedding}
As illustrated in Fig.~\ref{fig:pipeline}, our method takes a triangle surface mesh as input and computes the mapping that isometrically embeds the mesh vertices to a high dimensional (high-d) Euclidean space. The outputs are the vertex-wise coordinates of this high-d space. In the following subsections, we discuss the main technical components of our neural high-d Euclidean embedding method in detail: high-d Euclidean embedding loss and the network architecture. Due to the page limit, data generation for embedding and meshing is given in Section A of Supplemental Document.

\subsection{High-D Euclidean Embedding Loss}
Anisotropy represents how distances and angles are distorted, which can be measured by the dot product in geometry. For a metric defined over the surface domain $\Omega$, $M(.)\in \mathbb{R}^3$, at a given point $\mathbf{x}\in \Omega$, the dot product between two vectors $\mathbf{a}$ and $\mathbf{b}$ that starting from $\mathbf{x}$ is denoted by $\langle \mathbf{a},\mathbf{b}\rangle_{{M}(\mathbf{x})}$, which is defined over the tangent space of the surface: $\langle \mathbf{a},\mathbf{b} \rangle_{{M}(\mathbf{x})} = \mathbf{a}^t \mathbf{M}(\mathbf{x})\mathbf{b}$.

Our goal is to construct a high-d space $\mathbb{R}^{\overline{m}}$, in which the surface can be embedded with Euclidean metric. Inspired by~\cite{nash1956imbedding}, considering the high-d corresponding vectors $\overline{\mathbf{a}}$ and $\overline{\mathbf{b}}$, the dot product between $\overline{\mathbf{a}}$ and $\overline{\mathbf{b}}$ should introduce the same meaning of the dot product between $\mathbf{a}$ and $\mathbf{b}$ under the given metric, namely:\vspace{-2mm}
\begin{align}
    \langle \overline{\mathbf{a}},\overline{\mathbf{b}}\rangle = \langle \mathbf{a},\mathbf{b} \rangle_{M(\mathbf{x})}. \vspace{-4mm}
    \label{eq3}
\end{align}
Equation (\ref{eq3}) can be proved by the pullback metric. Considering the smooth mapping $\phi$ between $\mathbb{R}^3$ and $\mathbb{R}^{\overline{m}}$ at the point $\mathbf{x}$, the transformation between the vector in $\mathbb{R}^3$ and its correspondence in $\mathbb{R}^{\overline{m}}$ can be defined by the Jacobian matrix $\mathbf{J}(\mathbf{x})$ of $\phi$, in essence, $\overline{\mathbf{a}} = \mathbf{J}(\mathbf{x})\mathbf{a}$ and $\overline{\mathbf{b}} = \mathbf{J}(\mathbf{x})\mathbf{b}$. Therefore, $\langle \overline{\mathbf{a}},\overline{\mathbf{b}}\rangle = \mathbf{a}^t \mathbf{J}(\mathbf{x})^t\mathbf{J}(\mathbf{x})\mathbf{b} = \langle \mathbf{a},\mathbf{b}\rangle_{M(\mathbf{x})}$, where $\mathbf{M}(\mathbf{x}) = \mathbf{J}(\mathbf{x})^t\mathbf{J}(\mathbf{x})$.

\begin{wrapfigure}{r}{0.15\textwidth}\hspace{-5mm}
    \includegraphics[width=0.15\textwidth]{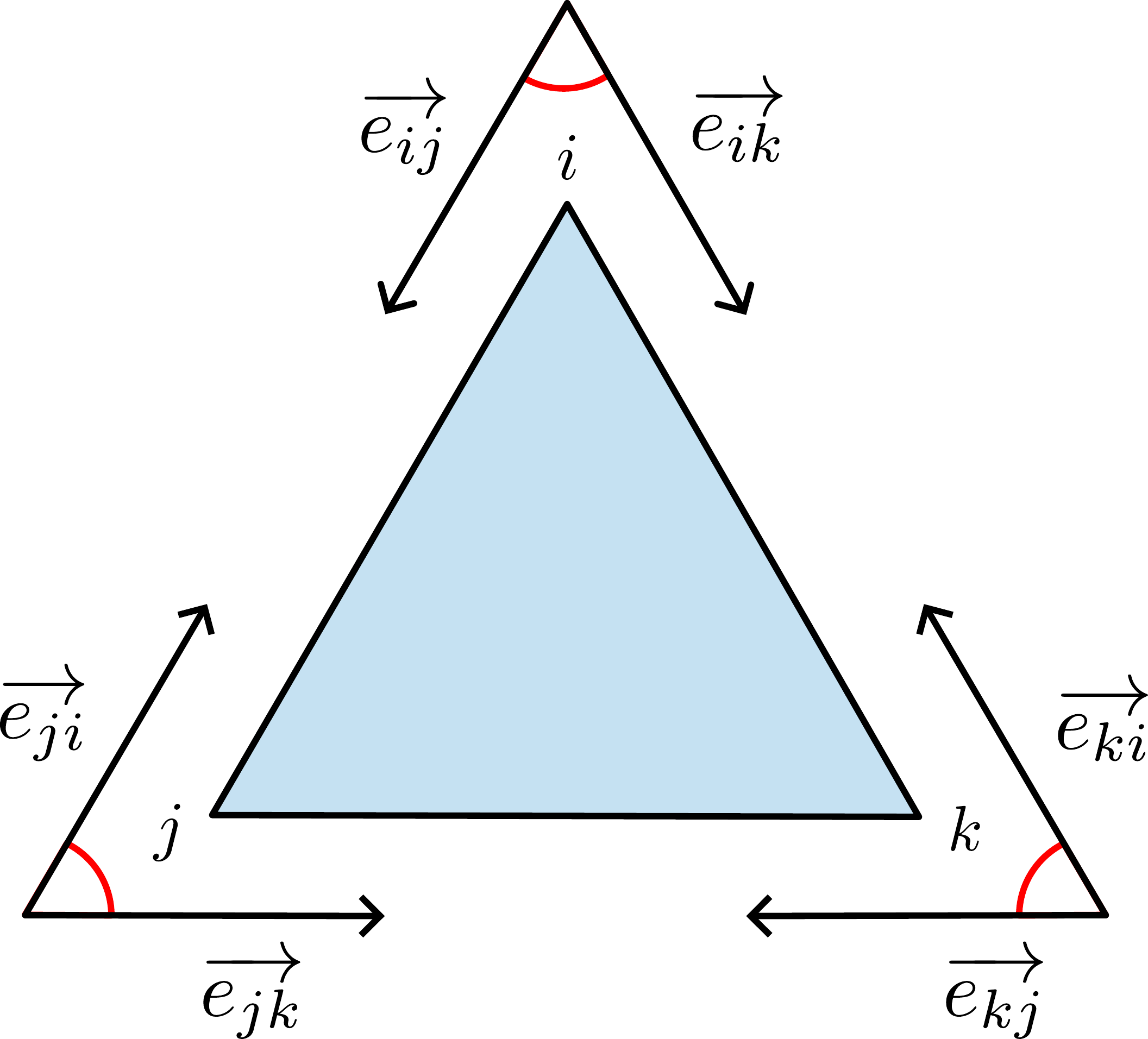}\vspace{-5mm}
\end{wrapfigure}
Our designed loss for neural Euclidean embedding leverages the advantage that the dot product between the vertices in $\mathbb{R}^{\overline{m}}$ has already contained their metrics in $\mathbb{R}^3$. Additionally, considering the discrete nature of mesh, a metric can be interpreted as influencing a piecewise linear region and distorted the dot product defined within the region. We can view those regions as the simplices that constitute the entire manifold surface. Therefore, we define our loss on the dot product between two edge vectors defined on an internal face angle via Mean Squared Error:
\begin{align*}
    \mathcal{L}_{dot} = \frac{1}{3|\mathbb{F}|}\sum_{\mathcal{F} \in \mathbb{F}}\big|\langle \xi(\mathbf{e}_{ij}), \xi(\mathbf{e}_{ik}) \rangle -\langle \overline{\mathbf{e}}_{ij}, \overline{\mathbf{e}}_{ik} \rangle \big|^2 \\
    \hspace{5mm}+ \big|\langle \xi(\mathbf{e}_{jk}),
    \xi(\mathbf{e}_{ji}) \rangle -\langle \overline{\mathbf{e}}_{jk}, \overline{\mathbf{e}}_{ji} \rangle \big|^2 \vspace{-6mm}
\end{align*}
\begin{equation}
\hspace{21mm} + \big|\langle \xi(\mathbf{e}_{ki}), \xi(\mathbf{e}_{kj}) \rangle -\langle \overline{\mathbf{e}}_{ki}, \overline{\mathbf{e}}_{kj} \rangle \big|^2,
\end{equation}
where $\xi(\mathbf{e}_{ij}) = f({\mathbf{v}}_j) - f({\mathbf{v}}_i)$, $f({\mathbf{v}}_i)$ and $f({\mathbf{v}}_j)$ are predicted high-d edge vector and vertex coordinates, which are learned from the input 3D coordinates; $\overline{\mathbf{e}}_{ij} = \overline{\mathbf{v}}_j - \overline{\mathbf{v}}_i$, $\overline{\mathbf{v}}_i$ and $\overline{\mathbf{v}}_j$ are the ground truth high-d edge vector and vertex coordinates. The ground truth data is geneated based on $\text{SIFHDE}^2$~\cite{Embedding2018} (refer Section A of Supplemental Document for details). Similar definitions are made for other edge vectors and vertices. So, each triangle face $\mathcal{F}$ has three dot product loss terms. 

The dot product loss can well align the metric distortion of the learned Euclidean embedding surface with the ground truth, but neural network tends to overfit the local distortion. To solve this problem, we add a Laplacian loss term as regularization:
\begin{align}
    \mathcal{L}_{lap} = \sum_{i\in \mathcal{V}} \bigg\|\frac{\sum_{j\in N(i)} (f(\mathbf{v}_i)-f(\mathbf{v}_j))}{|N(i)|}
    - \frac{\sum_{j\in N(i)} (\overline{\mathbf{v}}_i-\overline{\mathbf{v}}_j)}{|N(i)|}\bigg\|^2_2,
\end{align}
where ${N}(i)$ is the set of one-ring neighbors of vertex $i$.

Thus, our total loss is defined as the weighted sum of two losses:
\begin{align}
    \mathcal{L} = \mathcal{L}_{dot} + w_{lap}\mathcal{L}_{lap},
\end{align}
where $w_{lap} = 0.1 $ is based on extensive experiments. The analysis of $w_{lap}$ is discussed in Section F.2 of Supplemental Document.

\subsection{High-D Euclidean Embedding Network}
\label{sec:embed_net}
In order to take advantage of the connectivity of mesh, we utilize the graph neural network (GNN) to learn the high-d embedding. Specifically, given an surface mesh $\Omega \in \mathbb{R}^3$, its vertices $\mathcal{V}$ and edges $\mathcal{E}$ naturally define an undirected graph. We introduce the details about our network design in this section.

\subsubsection{Graph Convolution} The main purpose of graph convolution layer is to learn how to aggregate feature information from a node's local neighborhood and how to update its own feature. We employ the updating scheme based on~\cite{hamilton2018inductive} for our high-d embedding task. The convolution layer in our network follows:\vspace{-1mm}
\begin{align}
    f^{k+1}_i = \mathcal{W}^{k+1}_0\Big(f^{k}_i,\max_{j\in \mathcal{N}(i)}\mathcal{W}^{k+1}_1\mathcal{A}\big(f_i^{k},f_j^{k}\big)\Big),\vspace{-2mm}
\end{align}
where $\mathcal{W}_0$ and $\mathcal{W}_1$ are learnable parameters. $f^{k+1}$ and $f^{k}$ are the feature vectors on vertex $i$ before and after the convolution. $\mathcal{N}(i)$ is the set of one-ring neighbors of vertex $i$. $\mathcal{A}$ is a differentiable function to process feature information through the edge between vertex $i$ and $j$.

In pursuit of tailoring the convolution layer for learning high-d embedding, we want the learned extended coordinates to compensate the distortion / deformation followed by the metric. Therefore, we explicitly incorporate the direction and distance between each vertex and its neighbors into the aggregated feature information. $\mathcal{A}$ is accordingly defined as:\vspace{-1mm}
\begin{align}\label{conv_feature}
    \mathcal{A}\Big(f^k_i,f^k_j\Big) = \Big[f^k_j , f^k_j - f^k_i ,\big\|f^k_j - f^k_i\big\|_2\Big].\vspace{-2mm}
\end{align}

\subsubsection{Network Architecture} We construct a Graph U-Net~\cite{gao2019graph} with our tailored message passing paradigm. The network takes a 3D surface mesh as input, and embeds each vertex into the high-d space. The output keeps the same mesh connectivity as input. Fig.~\ref{fig:network} illustrates the architecture of the proposed high-d Euclidean embedding network. We build our residual blocks by two layers of graph convolution with skip connection followed by a batch normalization~\cite{ioffe2015batch} and a Leaky ReLU activation~\cite{maas2013rectifier}. Each down-sampling block is constituted by a residual block and one layer of Top-K pooling~\cite{gao2019graph}. Each up-sampling block is constituted by a residual block and one layer of interpolation. We employ five down-sampling blocks and five up-sampling layers. The network input is composed with six channels for each vertices, including vertex coordinates and normals. The output has five channels, and we concatenate the original 3D coordinates in front of the output to form a self-intersection free 8D embedding coordinates as in~\cite{Embedding2018}.\vspace{-2mm}
\begin{figure}[t!]
    \centering
    \includegraphics[height=6cm]{fig/neural_network.pdf}\vspace{-2mm}
    \caption{The architecture of the proposed high-d Euclidean embedding network. Each residual block combines two graph convolution layers with skip-connection. Each convolution layer is followed by a normalization layer and an activation layer except the output layer. The numbers represent the feature dimensions of each network layer. The graph convolution computation and neighboring feature aggregation are illustrated in detail.}\vspace{-2mm}
    \label{fig:network}
\end{figure}

\subsubsection{Training Data Augmentation} The diversity in shape meshes within our training set introduces biases in various aspects, particularly in the stretching direction and the distribution of stretching ratio across different parts of the mesh. To mitigate these issues and ensure a more robust and generalized model performance, we augment the training data by rotation (by $\pi/2$ around three Euclidean axes), and mirroring (according to $xy$, $xz$, and $yz$ planes). These two augmentation types reorient the stretching direction and redistribute the stretching ratio according to vertex coordinates which ensure the anisotropy property is represented more uniformly.

%% file: section/4_normal_metrci_cvt.tex
\section{Feature-Sensitive Anisotropic Meshing}
Remeshing 3D surfaces with features is a challenging problem, not to mention anisotropic remeshing. Existing anisotropic remeshing approaches~\cite{Embedding2018,fu2014anisotropic} let the user specify which edges correspond to features, and use constrained optimization to sample them properly. This is tedious in practical for 3D surface shapes with different features. In this work, after mapping the surface $\Omega\in \mathbb{R}^3$ into the neural high-d Euclidean space, the follow-up task is to isotropically remesh this high-d embedded surface $\overline{\Omega}\in \mathbb{R}^{\overline{m}}$. 
Inspired by~\cite{levy2010LpCVT}, we re-design the normal metric from 3D to high-d space, and then formulate a new optimization energy function for high-d CVT ($d = 8$ in our case) with quadratic normal metric to automatically preserve features and anisotropy that exhibit in the original 3D shapes.

\subsection{Normal Metric in High-D}
The original normal metric $\mathbf{M}^{\mathcal{T}}$ associated with facet $\mathcal{T}$ in 3D is defined as follows~\cite{levy2010LpCVT}: $\mathbf{M}^{\mathcal{T}} = (s-1)\begin{pmatrix}
         \mathbf{N}^{\mathcal{T}}_x[\mathbf{N}^{\mathcal{T}}]^t\\
         \mathbf{N}^{\mathcal{T}}_y[\mathbf{N}^{\mathcal{T}}]^t\\
         \mathbf{N}^{\mathcal{T}}_z[\mathbf{N}^{\mathcal{T}}]^t\\
     \end{pmatrix}+\mathbf{I}_{3\times3}$, 
where $\mathbf{N}^{\mathcal{T}}$ denotes the unit normal of facet ${\mathcal{T}}$ in 3D and $s$ is a factor to emphasize the normal metric ($s = 7$ in the experiments). 

In our high-d embedded surface, since we only focus on the features that are identified in the original 3D surface, we can extend $\mathbf{M}^{\mathcal{T}}$ by padding $1s$ on the diagonal of $\mathbf{M}^{\mathcal{T}}$ and $0s$ otherwise to define the normal metric $\overline{\mathbf{M}}^{\mathcal{T}}$ in the high-d space $\mathbb{R}^{\overline{m}}$:
\begin{align}
    \overline{\mathbf{M}}^{\mathcal{T}} = \begin{pmatrix}
        \mathbf{M}^{\mathcal{T}} & 0\\
        0 & \mathbf{I}_{hd}\\
    \end{pmatrix},\vspace{-2mm}
\end{align}
where $\mathbf{I}_{hd}$ is an $(\overline{m}-3)\times (\overline{m}-3)$ identity matrix. $\overline{\mathbf{M}}^{\mathcal{T}}$ is an orthogonal matrix, which does not affect the Euclidean distance calculated from the high-d space. In this way, the proposed high-d quadratic normal metric can penalize the remeshing vertices that are far away from the tangent plane of the high-d embedding. On a feature edge, the combined effects of the normal metrics of both facets incident to a feature edge tend to attract remeshing vertices onto such edge.
 
\subsection{High-D Normal Metric CVT}\label{sec:NMCVT}
Inspired~\cite{levy2010LpCVT} (in 3D), we define the combinatorial structure and algebraic structure of high-d normal metric CVT. To compute high-d CVT is to minimize the energy function $E_{hd}$. We use gradient-based optimization method which needs to evaluate $E_{hd}$ and its gradient $\nabla E_{hd}$. For each iteration, the high-d embedding surface is first decomposed into a set of simplicial triangle facets through a differentiable clipping algorithm (in Section B of Supplemental Document). By doing so, the expression of $E_{hd}$ can be simply evaluated. After having the combinatorial representation, the value of $E_{hd}$ is obtained by the closed form and $\nabla E_{hd}$ can be obtained by applying chain rule and reverse-mode differentiation (in Section C of Supplemental Document).

The objective function of feature-sensitive high-d CVT is defined by adding a high-d normal metric $\overline{\mathbf{M}}^{\mathcal{T}}$ into CVT energy in high-d:\vspace{0mm}
\begin{align}
\label{eq:hdcvt_energy}
    E_{hd}(\mathbf{\overline{X}}) &= \sum_i \underset{\overline{\Omega}_i\cap\mathcal{S}}{\int} \Big\|\overline{\mathbf{M}}^{\mathcal{T}}[\mathbf{\overline{y}}-\mathbf{\overline{x}}_i]\Big\|^2_2 d\mathbf{\overline{y}},\vspace{-2mm}
\end{align}
where $\mathbf{\overline{x}}_i \in \mathbf{\overline{X}}$ are the Voronoi cell sites. For RVD on the high-d embedded surface mesh, Voronoi cells are discretized to a set of facet triangles, denoted as $\mathcal{T}(\mathbf{\overline{x}}_0,\mathbf{{C}}_1,\mathbf{{C}}_2,\mathbf{{C}}_3)$. $\mathbf{{C}}_1,\mathbf{{C}}_2,\mathbf{{C}}_3$ are three different configurations of the vertices on these facets triangles, which need to be treated differently in order to have the accurate gradient. Details are given in Section B of Supplemental Document. So $\overline{\Omega}$ is the sum of the area of all the facets. A high-d CVT is a stable and critical point of $E_{hd}$ during the optimization. 

As stated in~\cite{levy2010LpCVT}, given the gradient of standard CVT energy $E$, $\nabla E = 2m_i(\mathbf{x}_i-\mathbf{g}_i)$, one cannot simply replace the centroid $\mathbf{g}_i$ and mass ${m}_i$ with their anisotropic counterparts to get the gradient of high-d CVT energy $E_{hd}$, since the anisotropy varies between two adjacent cells. The closed form for gradient of high-d CVT with normal metric over an integration simplex $\mathcal{T}$ is:
 \begin{equation}
 \label{eq:discretize1}
     \frac{dE^{\mathcal{T}}_{hd}(\mathbf{\overline{x}}_0,\mathbf{{C}}_1,\mathbf{{C}}_2,\mathbf{{C}}_3)}{d\mathbf{\overline{X}}} = 
     \frac{dE^{\mathcal{T}}_{hd}}{d\mathbf{\overline{x}}_0}+ \sum_{i=1,2,3} \frac{dE^{\mathcal{T}}_{hd}}{d\mathbf{{C}}_i}\frac{d\mathbf{{C}}_i}{d\mathbf{\overline{X}}},
 \end{equation}
where $\mathbf{\overline{X}}$ is the site point of the Voronoi cell. $\mathcal{T}(\mathbf{{C}}_1,\mathbf{{C}}_2,\mathbf{{C}}_3)$ is one of the facet triangles that compose the restricted Voronoi cell in the high-d embedding space.

\subsection{Auto Differentiation for High-D Normal Metric CVT}
Automatic differentiation is a computational technique that leverages the chain rule of calculus. Instead of computing gradients from an explicit formula, automatic differentiation computes gradient starting from the function value, and tracing backwards along how the function value has been computed; and leverages the chain rule to compute the gradient. 

From Equation~(\ref{eq:discretize1}), the calculation of gradient can be separated to two parts: $\frac{dE^{\mathcal{T}}_{hd}}{d\mathbf{C}_i}$ and $\frac{d\mathbf{C}_i}{d\mathbf{\overline{X}}}$, and the total gradient can be assembled from these two expressions. In this subsection, we introduce how to calculate $E^{\mathcal{T}}_{hd}$ and $\mathbf{C}_i$ in the forward pass in order to get the correct gradient $\frac{dE^{\mathcal{T}}_{hd}}{d\mathbf{C}_i}$ and $\frac{d\mathbf{C}_i}{d\mathbf{\overline{X}}}$ in the reverse pass.

For $E^{\mathcal{T}}_{hd}$, it can be discretized onto the triangle facets that compose the restricted Voronoi cell, and the expression for this discretization can be shortly expressed as:
\begin{equation}
    E^{\mathcal{T}}_{hd}=|{\mathcal{T}}|F^{\mathcal{T}}_{hd},
\end{equation}
where $|{\mathcal{T}}|$ denotes the area of current triangle facet, referring Section A in~\cite{levy2010LpCVT}'s Appendix for details. 


It is noted that our computation is more complicated and challenging than~\cite{levy2013variational}, which also uses Heron's formula to calculate the area of triangle in 6D, but without having the metric in their CVT energy function. So, their 6D CVT optimization does not require to compute the gradient of Heron's formula. Conversely, for our high-d normal metric CVT, $|\mathcal{T}|$ becomes dependent regards to $E^T_{hd}$:\vspace{-2mm}
\begin{equation}\label{eq:dfdc}
    \frac{dE^{\mathcal{T}}_{hd}}{d\mathbf{C}_i} = (\frac{dF^{\mathcal{T}}_{hd}}{d\mathbf{U}_i}|\mathcal{T}|+\frac{d|\mathcal{T}|}{d\mathbf{U}_i}F^{\mathcal{T}}_{hd})\mathbf{\overline{M}}^{\mathcal{T}},
\end{equation}
where $\mathbf{U}_i$ is defined in Equation (3) of Supplemental Document. The detailed computations of automatic differentiation on $\frac{dE^{\mathcal{T}}_{hd}}{d\mathbf{C}_i}$ and the derivative of $\mathbf{C}_i$ are provided in Sections D.1 and D.2 of Supplemental Document, respectively.



\subsection{Restricted Voronoi Diagram and Mesh Generation}
Following the optimization of the CVT in high-d space, the barycentric coordinates of each site point can be utilized to back-project the RVD from the high-d embedding space onto the original three-dimensional space. This process results in the generation of the final anisotropic RVD and dual mesh. We leverage the advantage of Geogram~\cite{levy2015geogram}, which computes the RVD using filtered geometric predicates and symbolic perturbation to resolve degeneracies~\cite{levy_PCK}. In general, there is a possibility that when generating a mesh using dual mesh of RVD, i.e., the associated RDT, inverted elements may occur upon back-projection to 3D space. Our implementation addresses this issue by inserting additional points using a provably terminating algorithm~\cite{rouxel2016discretized} whenever such an inverted element is detected.\vspace{-1mm}

%% file: section/5_experiment.tex
\section{Experimental Results}
\subsection{Datasets}
\label{data_preparation}
In this work, the evaluations and applications mainly focus on better and faster approximating shapes by generating the anisotropic surface meshes on a large scale.\vspace{-1mm}

\subsubsection{Benchmark} 
In all of our experiments, we train our network on the selected models from Thingi10k dataset~\cite{Zhou:2016:MASG}. We select 240 meshes (after applying the data augmentation strategy proposed in Section~\ref{sec:embed_net}, there are 2,400 meshes in our experiments) for training and 280 meshes for testing. Meshes are selected to ensure that our neural network can effectively learn curvature-related information. In the dataset, we only exclude surface models predominantly composed of planar or zero mean curvature regions where there are no anisotropic / curvature properties at all. 
Majority of meshes in our current training dataset contain planar regions. This is already sufficient for our NASM to learn how to effectively apply isotropic distribution in high-d spaces to planar regions as demonstrated in our results. We use the surface meshes that are generated by TetWild~\cite{Hu:2018} and normalize them to [-1, 1] to make meshes with evenly distributed vertices and valid faces for building the ground truth data and evaluation use.\vspace{-1mm}

\subsubsection{Generalization} 
To validate the generalization ability, we further evaluate our method on two large-scale unseen datasets with different types of meshes. There are 154 garment mesh models selected from Multi-Garment Net (MGN)~\cite{bhatnagar2019mgn}, a dataset of clothes with open boundary, including 96 pants and 58 tops. Another dataset is 3D human FAUST~\cite{bogo2014faust} with 200 human scan models of 10 different subjects in 30 different poses. We test these two datasets by using the same network parameters trained on Thingi10k training set without any fine-tuning.

\subsection{Implementation Details}
For curvature metric generation, we first use Libigl~\cite{libigl} to calculate the curvatures and principal directions. The curvature metric is designed as follows: $\mathbf{M}=[\mathbf{v}_{min}, \mathbf{v}_{max}, \mathbf{n}]diag(1,(\frac{s_2}{s_1})^2,$\\
$1)[\mathbf{v}_{min},\mathbf{v}_{max},\mathbf{n}]^t$, where $\mathbf{v}_{min}$ and $\mathbf{v}_{max}$ are the directions of the principal curvatures, $\mathbf{n}$ is the unit surface normal. $s_1$ and $s_2$ are two stretching factors along principal directions. For surface anisotorpic task, $s_1=\sqrt{|K_{min}|}$ and $s_2=\sqrt{|K_{max}|}$ where $K_{min}$ and $K_{max}$ are the principal curvatures. We set small thresholds to preserve $K_{min}$ and $K_{max}$ not vanishing. To obtain smooth stretching factors, we compose $\frac{s_2}{s_1}$ with the principal curvatures calculated by Libigl and then apply weighted average over the one-ring neighborhood.

For ground truth high-d embedding generation, we implement the process in Section A of Supplemental Document by using Eigen~\cite{eigenweb} and leverage MUMPS~\cite{amestoy2001fully} sparse solver to accelerate the computation on AMD processor.

For the neural network development, we use PyTorch Geometric~\cite{Fey/Lenssen/2019} to build the network and employ AdamW\\
~\cite{loshchilov2018decoupled} optimizer with a learning rate 0.01 for training. We train our network for 600 epochs with batch size 4, and the learning rate is halved every 100 epochs.

For the high-d normal metric CVT development, we use Geogram~\cite{levy2015geogram} for restricted Voronoi diagram (RVD) calculation, Stan Math Library~\cite{carpenter2015stan} for auto differentiation gradient computation, and L-BFGS~\cite{liu1989limited} for CVT optimization. \textit{The source code of our framework and data will be publicly released after acceptance.}

\subsection{Evaluation Metrics}
Besides the qualitative visualization measurement, the quantitative evaluation for anisotropic mesh result includes: surface accuracy, mesh quality, and computational time. 

\subsubsection{Surface Accuracy}
For the evaluation on surface mesh accuracy, we use Chamfer Distance (CD), F-Score (F1), normal consistency (NC), and Hausdorff Distance (HD). To evaluate the ability of preserving sharp features, following PoNQ~\cite{maruani2024ponq}, we use Edge Chamfer Distance (ECD) and Edge F-score (EF1).

\subsubsection{Mesh Quality}
To measure the anisotropic mesh quality, for each triangle $\triangle_{abc}$ in the final mesh, we use its approximated metric $\mathbf{Q}(\triangle_{abc}) = (\mathbf{Q}(\mathbf{x}_a) + \mathbf{Q}(\mathbf{x}_b) + \mathbf{Q}(\mathbf{x}_c))/3$, where $\mathbf{Q}(\cdot) = \sqrt{\mathbf{M}(\cdot)}$, to affine-transform it from the original anisotropic space into the Euclidean space. After that, we employ the isotropic triangular criteria~\cite{frey1999surface}, to evaluate the quality of generated anisotropic triangular mesh, as suggested by~\cite{Particle2013,fu2014anisotropic}. The quality of a triangle is measured by $G = 2 \sqrt{3} \frac{S}{ph}$, where $S$ is the triangle area, $p$ is its half-perimeter, and $h$ is the length of its longest edge. $G_{avg}$ is the average qualities of all triangles. 

\subsubsection{Timing}
One of our main advantages is the efficiency of computing the high-d embedding. In Tab.~\ref{tab:testset}, we report the inference time $T_{em}$ of the neural high-d embedding and computational time of mesh $T_{me}$ on high-d normal metric CVT or high-d CVT, for the Thingi10k testing set. Furthermore, we report the inference time of our method with respect to the number of input mesh vertices for all three testing datasets in Section F.1 of Supplemental Document. The timings of our method and $\text{SIFHDE}^2$~\cite{Embedding2018} are collected from 3.8 GHz AMD Ryzen 3960x processor and an NVIDIA GeForce RTX 3090 GPU with 24GB GDDR6X. 
\subsection{Results on Surfaces without Sharp Features}
\begin{figure*}
\centering
    \includegraphics[width=0.98\textwidth]{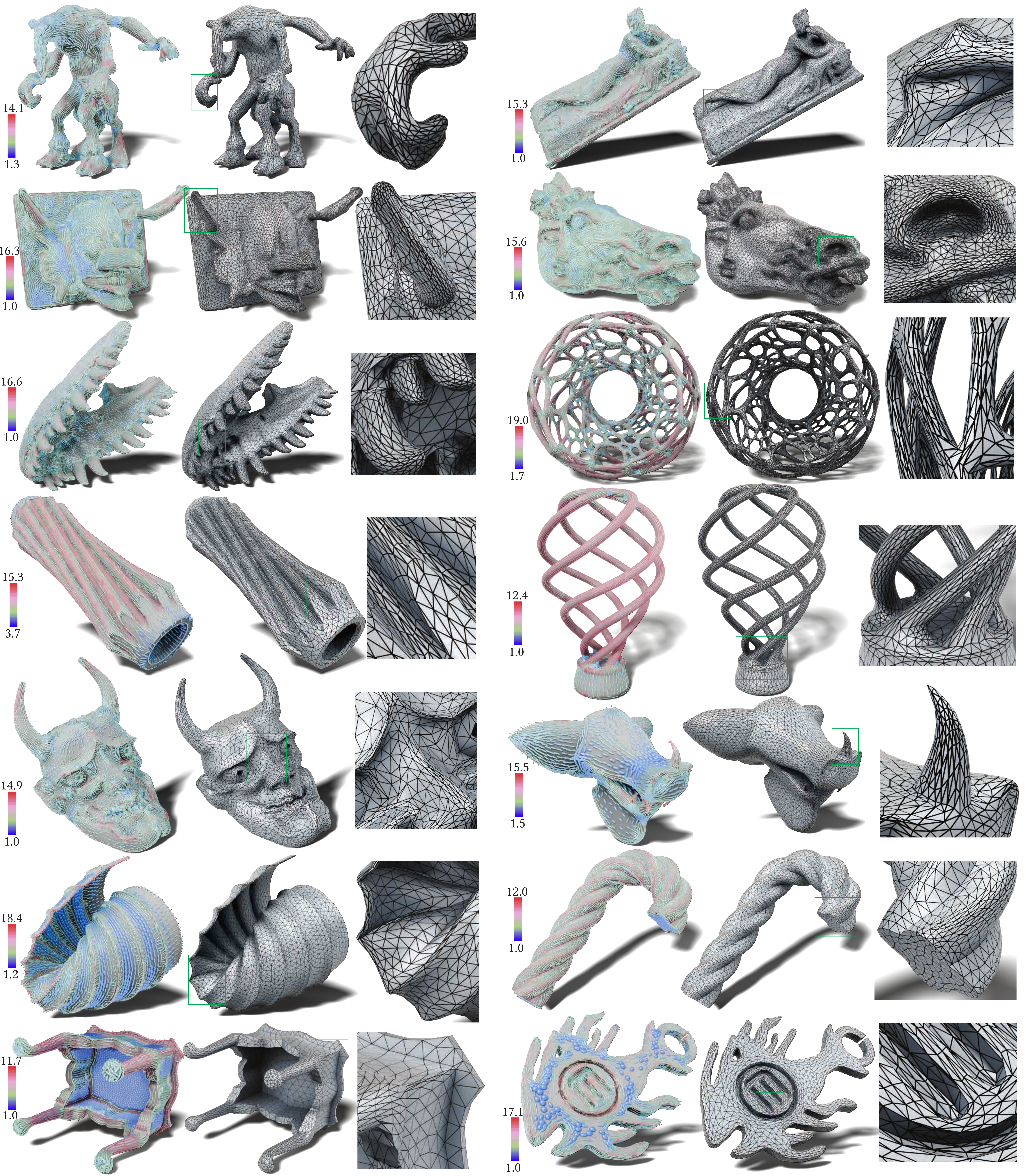}
    \caption{Our anisotropic surface meshing results on smooth surfaces (top three rows) and surfaces with sharp or weak features (bottom four rows). (left to right: curvature tensors with corresponding stretching ratios denoted in colors, anisotropic meshing, and a zoom-in illustration). The files' IDs are provided from Thingi10k dataset. Left column: $133077,76778,46461,741525,81589,87688,51015$; Right column: $72870,68380,61258,39086,40992,107910,75989$.}
    \label{fig:meshing_results}
\end{figure*}

\begin{figure*}\label{comp_sifhd}
\centering
    \includegraphics[width=0.92\linewidth]{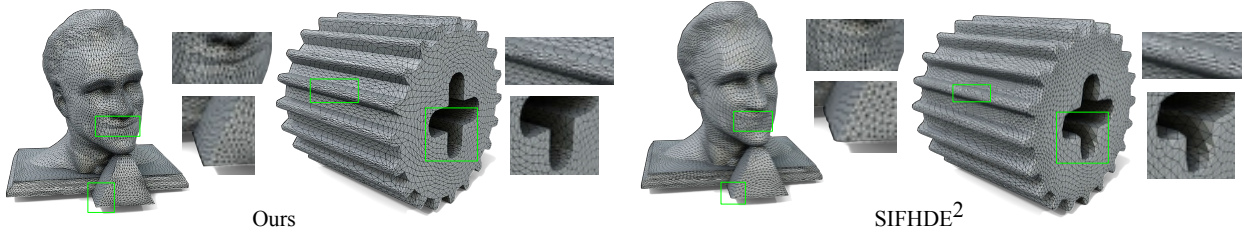}\vspace{-2mm}
    \caption{Comparison between our method and $\text{SIFHDE}^2$~\cite{Embedding2018} on models from Thingi10k dataset with the same number of vertices.}\vspace{-1mm}
    \label{fig:compare_embed}
\end{figure*}

We first evaluate our NASM method on the testing set with 280 mesh models selected from Thingi10k as shown in Fig.~\ref{fig:meshing_results}, and compare the result to $\text{SIFHDE}^2$~\cite{Embedding2018}. The original version of $\text{SIFHDE}^2$ heavily relies on the smoothness of input metric field, and we conduct the comparison experiments on our improved version (in Section A of Supplemental Document), for both the surface accuracy and anisotropic mesh quality measurement. The improved version of $\text{SIFHDE}^2$ takes surface mesh and its corresponding metric field as input. We use the same process to generate high-d embedding for our dataset preparation (in Section~\ref{data_preparation}). In their paper~\cite{Embedding2018}, they used high-d version of particle-based method~\cite{Particle2013}. However, the definition for inter-particle energy and force are defined on Euclidean distance, not on the surface manifold. For this reason, we leverage the high-d CVT from the meshing process, which uses RVD during the optimization. It can better to approximate the geodesic distance. 
The quantitative results between our method and $\text{SIFHDE}^2$ are reported in Tab.~\ref{tab:testset}. Our method outperforms $\text{SIFHDE}^2$ in both surface accuracy, anisotropic mesh quality, as well as computational time with about 1,500$\times$ speedup. Fig.~\ref{fig:compare_embed} shows the qualitative comparison with $\text{SIFHDE}^2$ method. Our method can better represent the local geometry changes, such as the zoom-in regions for selected models from Thingi10k dataset. 
\begin{table}
\caption{Quantitative comparison with our NASM, NASM w/o high-d normal metric CVT, and $\text{SIFHDE}^2$ method~\cite{Embedding2018} on 80 models selected from Thingi10k dataset, including surfaces without and with sharp and weak features. All the evaluation metrics are average values of all models from the dataset. The best results are highlighted in bold per different $\#V_{out}$. Note: $\#V_{in}$ and $\#V_{out}$ are the average numbers of vertices of all input and output meshes, `Stretch' is the average anisotropic stretching ratios of all models, CD ($\times 10^5$), HD ($\times 10^2$), ECD ($\times 10^2$), $T_{em}$ (s), $T_{me}$ (s).}\vspace{-3mm}
\label{tab:testset}
\begin{center}
\resizebox{0.47\textwidth}{!}{%
    \begin{tabular}{lllllllllllll}
    \toprule
     Method &  $\#V_{in}$ & $\#V_{out}$ & Stretch & CD $\downarrow$ & F1 $\uparrow$ & NC $\uparrow$ & HD $\downarrow$ & ECD  $\downarrow$ & EF1 $\uparrow$ & $T_{em}$  $\downarrow$ & $G_{avg}$ $\uparrow$ & $T_{me}$  $\downarrow$ \\
    \midrule
    NASM & 5,982 & 5,982  & 12.736 & \textbf{0.709} & \textbf{0.978} & \textbf{0.993} & \textbf{0.725} & \textbf{0.066} & \textbf{0.897} & \textbf{0.029} & 0.745 & 14.022 \\
    & 5,982 & 3,590 & 12.736& \textbf{0.720} & \textbf{0.978} & \textbf{0.991} & \textbf{0.779} & \textbf{0.086} & \textbf{0.850} & \textbf{0.029} & 0.748 & 7.366 \\
    & 5,982 & 1,202  & 12.736& \textbf{0.882} & \textbf{0.967} & \textbf{0.984} & \textbf{1.127} & \textbf{0.148} & \textbf{0.676} & \textbf{0.029} & 0.744 & 3.499 \\
    &5,982  & 608  & 12.736& \textbf{1.538} & \textbf{0.928} & \textbf{0.974} & \textbf{1.774} & \textbf{0.207} & \textbf{0.501} & \textbf{0.029} & 0.732 & \textbf{2.092} \\
    \midrule
    NASM  & 5,982 & 5,982 & 12.736 & 0.779 & 0.972 & 0.989 & 0.882 & 0.137 & 0.687 & \textbf{0.029} & \textbf{0.758} & 3.780 \\
    w/o NM CVT &5,982  & 3,590  &12.736 & 0.875 & 0.963 & 0.987 & 1.055 & 0.155 & 0.571 & \textbf{0.029} & \textbf{0.764} & \textbf{3.354} \\  
     (w/ CVT)& 5,982 & 1,202  & 12.736& 1.684 & 0.905 & 0.975 & 1.613 & 0.215 & 0.290 & \textbf{0.029} & \textbf{0.776} & 3.127 \\
    & 5,982 & 608 &12.736 & 3.692 & 0.779 & 0.962 & 2.352 & 0.267 & 0.170 & \textbf{0.029} & \textbf{0.779} & 2.922 \\
     \midrule
    $\text{SIFHDE}^2$ & 5,982 & 5,982 & 12.736 & 0.808 & 0.969 & 0.988 & 0.949 & 0.146 & 0.612 & 49.25 & 0.729 & \textbf{3.718} \\
    & 5,982 & 3,590 & 12.736& 0.928 & 0.959 & 0.985 & 1.145 & 0.166 & 0.487 & 49.25 & 0.732 & 3.441 \\
    & 5,982 & 1,202 &12.736 & 1.878 & 0.893 & 0.975 & 1.848 & 0.222 & 0.249 & 49.25 & 0.734 & \textbf{3.005} \\
    & 5,982 & 608 &12.736 & 4.144 & 0.766 & 0.963 & 2.674 & 0.270 & 0.157 & 49.25 & 0.730 & 2.935 \\
    \bottomrule
    \end{tabular}%
}\vspace{-2mm}
\end{center}
\end{table}

\begin{figure*}
\centering
    \includegraphics[width=0.8\linewidth]{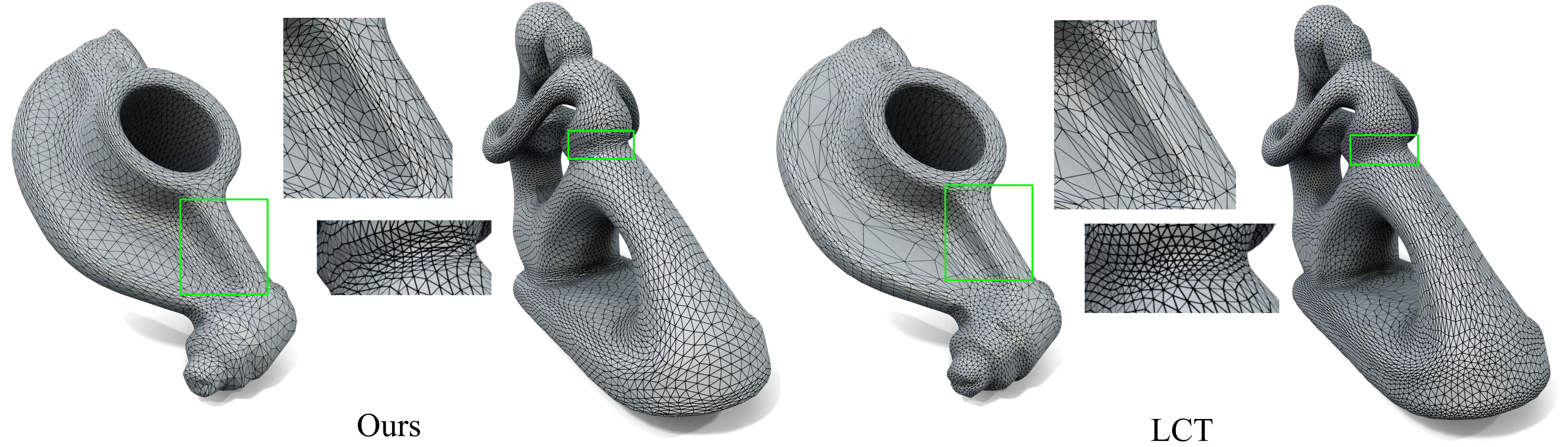}
    \vspace{-10pt}
    \caption{Comparison between our method and LCT method~\cite{fu2014anisotropic} on Rocker Arm and Fertility models with the same number of vertices.}\vspace{-2mm}
    \label{fig:compare_lct}
\end{figure*}

We show further comparative analysis and experiments with another state-of-the-art anisotropic surface meshing approach, Locally Convex Triangulation (LCT)~\cite{fu2014anisotropic} on Fertility and Rocker Arm models. From Fig.~\ref{fig:compare_lct}, in highlighted regions, it is clear to see that our method can capture the local curvatures more accurately. However, the stretching ratios of the mesh elements in LCT result are either too big or too small on some regions, since their method highly depends on the specific input curvature metric. The curvature metric computation is not stable and variant due to the different mesh discretizations. In Tab.~\ref{tab:comparison_LCT}, it shows that our method has better surface accuracy on CD, F1, NC, HD metrics. LCT is a bit better than ours on mesh quality $G_{avg}$. Due to lacking their curvature metrics, LCT's $G_{avg}$ values are copied from the original paper.
\begin{table}
\caption{Quantitative comparison with our NASM and LCT method~\cite{fu2014anisotropic} on Fertility and Rocker Arm models. The best results are highlighted in bold. Note: CD ($\times 10^5$) and HD ($\times 10^2$). Note: *Due to lacking their curvature metrics, LCT's $G_{avg}$ values are copied from their original paper.}\vspace{-3mm}
\label{tab:comparison_LCT}
\begin{center}
\resizebox{0.35\textwidth}{!}{%
    \begin{tabular}{lllllllllllll}
    \toprule
    Model  & Method & $\#V_{out}$ &CD $\downarrow$& F1 $\uparrow$ & NC$\uparrow$ & HD $\downarrow$ & $G_{avg}$ $\uparrow$ \\
    \midrule
    Rocker Arm  & LCT &5,550 & 0.625 & 0.995 & 0.994 & 0.597 & \textbf{0.86}*\\
             &NASM& 5,547 & \textbf{0.600} & \textbf{0.996} & \textbf{0.996} & \textbf{0.576} & 0.722 \\
    Fertility & LCT & 12,480& 0.584 & \textbf{0.996} & \textbf{0.996} & 0.603&\textbf{0.89}*\\
              & NASM & 12,475& \textbf{0.578} & \textbf{0.996} & \textbf{0.996} & \textbf{0.596} & 0.712 \\
    \bottomrule
    \end{tabular}%
}\vspace{-2mm}
\end{center}
\end{table}

\subsection{Results on Surfaces with Sharp Features}
\begin{figure*}
\centering
    \includegraphics[width=0.92\textwidth]{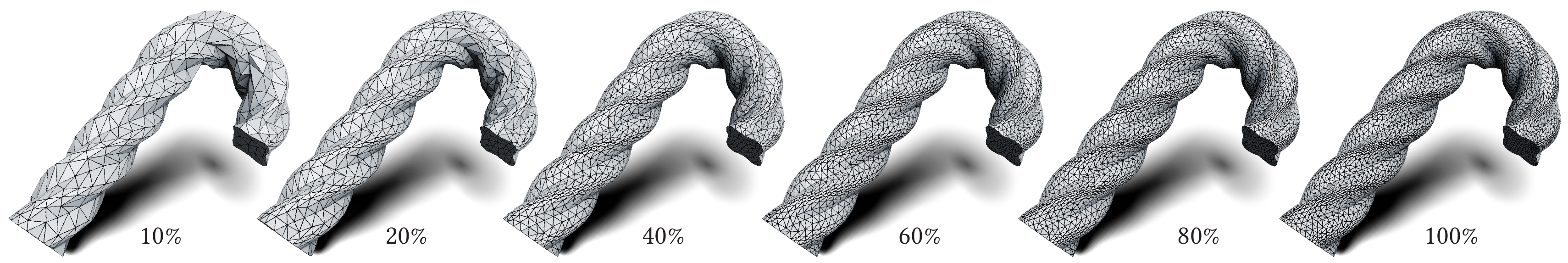}\vspace{-10pt}
    \caption{Anisotropic surface meshing results of our NASM method on different number of output vertices (from $10\%$ to $100\%$ of the user specified vertex numbers, i.e., 555, 1103, 2207, 3311, 4415, 5518). It is clear to see that even $10\%$ of vertices, our method can still well capture both the curvature metrics and features.}\vspace{-4mm}
    \label{fig:mul-res}
\end{figure*}

Another advantage of our NASM method is to automatically keep sharp features from the input mesh without any user's intervention as shown in Fig.~\ref{fig:meshing_results}. To show the robustness of our ability of feature preserving, we conduct the experiments on different level of resolutions for output meshes and compare the results with $\text{SIFHDE}^2$. We use $100\%$, $60\%$, $20\%$, and $10\%$ of vertex count from input meshes and make the quantitative report in Tab.~\ref{tab:testset} (a full version is included in Section E.1 of Supplemental Document). Our NASM outperforms $\text{SIFHDE}^2$ at all levels of resolution. It obtains better surface accuracy and anisotropic quality than $\text{SIFHDE}^2$. Fig.~\ref{fig:mul-res} shows one example on different resolutions for anisotropic meshes. This can be applied in the curvature-induced mesh simplifications.


Furthermore, we do an ablation study on normal metric CVT for our NASM by using a general CVT to generate the anisotropic mesh. The result is reported in Tab.~\ref{tab:testset}. We observe an increasing of anisotropic mesh quality when using CVT for meshing. Based on our analysis, it is actually a trade-off between anisotropy mesh quality and surface accuracy (feature preserving). Normal metric CVT can push the vertices towards sharp feature, which may distort the anisotropic metric direction and stretching ratio. So that the mesh quality evaluation may become lower. 

The qualitative comparison with NASM method, NASM without high-d normal metric CVT (with CVT), and $\text{SIFHDE}^2$ method is provided in Fig. 2 of Supplemental Document. Fig.~\ref{fig:rvd} shows anisotropic RVD results on some complicated surfaces from the Thingi10k dataset, which are computed by our high-d normal metric CVT optimization. Some additional RVD results are shown in Fig. 2 of Supplemental Document due to the page limit.\vspace{-2mm} 
\begin{figure*}
\centering
    \includegraphics[width=0.9\textwidth]{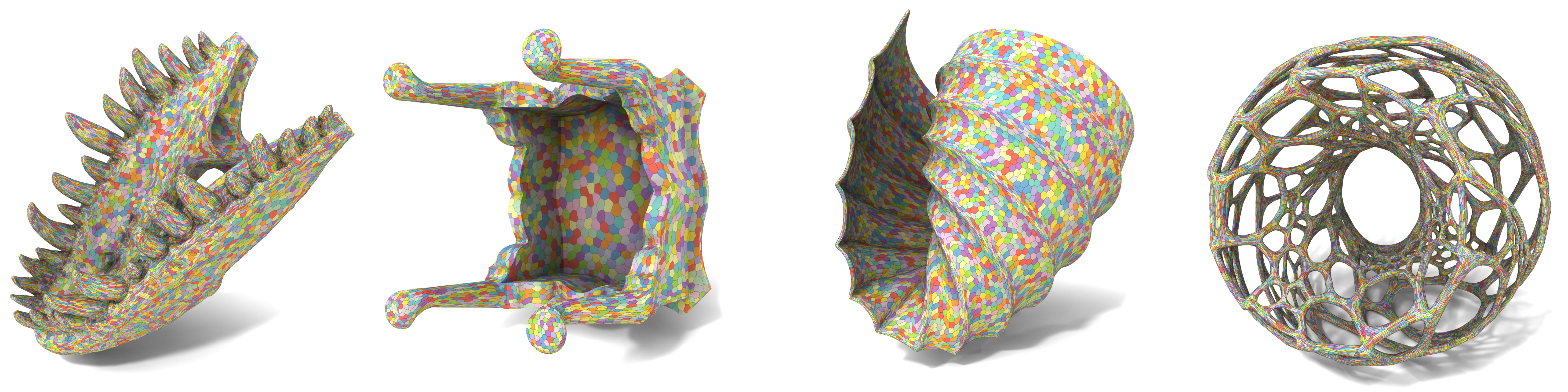}\vspace{-14pt}
    \caption{Anisotropic RVD results on some complicated surfaces from the Thingi10k dataset, which are computed by our high-d normal metric CVT optimization. The files' IDs are provided from Thingi10k dataset from left to right: $46461,51015,87688,61258$.}
    \label{fig:rvd}
\end{figure*}

\subsection{Results on Unseen Datasets}
\begin{figure*}
\centering
    \includegraphics[width=0.92\textwidth]{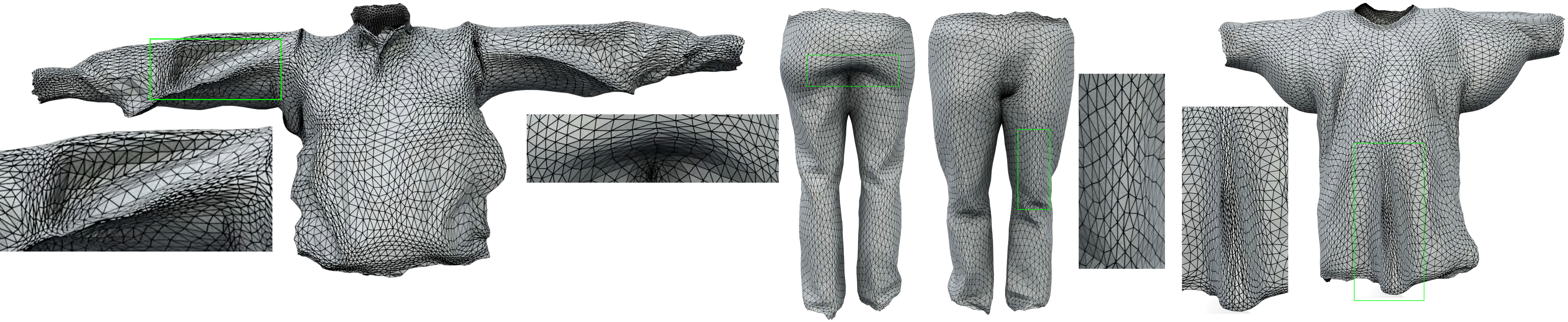}\vspace{-10pt}
    \caption{Anisotropic surface meshing results of our NASM method on an unseen testing dataset, e.g., MGN dataset. The examples of our results can well capture the open boundaries and detailed cloth wrinkles and folds from tops and pants models.}
    \label{fig:cloth}
\end{figure*}

To further demonstrate the robustness and extensibility of our method, we train our neural high-d embedding on Thingi10k dataset, and directly test it on 154 models (including 96 pants and 58 tops) from MGN dataset without fine-tuning the network parameters. It is noted that MGN models are quite different from the training mesh models. Fig.~\ref{fig:cloth} shows that our results can well capture open boundaries, detailed anisotropies and features around cloth wrinkles and folds from pants models. More visualization results and quantitative evaluations are provided in Section E.2 of Supplemental Document.

We also test our method on part of FAUST dataset with 200 human scan models of 10 different subjects in 30 different poses without fine-tuning the network parameters. Each scan is a high-resolution and non-watertight triangular mesh. We use QEM~\cite{garland1997surface} to reduce the mesh resolution to feed into the neural network. Both qualitative and quantitative evaluations are provided in Section E.3 of Supplemental Document. Our results show promising results to capture geometric anisotropies and features around hands, arms, legs, and thin clothes wrinkles on 3D human models.

\subsection{Ablation Study}
We provide quantitative and qualitative ablation studies in Tab.~\ref{tab:comparison_loss} and Fig. 9 of Supplemental Document on different loss functions, and data augmentation. The ablation study is performed on the Thingi10k dataset. Compared with L2 loss and Cosine loss, our proposed dot product loss is most effective on the anisotropic surface meshing. Our loss outperforms other losses on all the surface accuracy and mesh quality metrics. Visually, it is noted that the dot product loss can better recover the curvature metrics as well as better capture the geometric features. 
\begin{table}
\caption{Ablation study on the losses of our dot product, L2, Cos, and without data augmentation on 80 models from Thingi10k dataset. All the evaluation metrics are average values from the dataset. The best results are highlighted in bold. Note: CD ($\times 10^5$), HD ($\times 10^2$), ECD ($\times 10^2$).}\vspace{-3mm}
\label{tab:comparison_loss}
\begin{center}
\resizebox{0.47\textwidth}{!}{%
    \begin{tabular}{lllllllllllll}
    \toprule
    Loss / Method &$\#V_{in}$ & $\#V_{out}$ $\downarrow$ &CD $\downarrow$& F1 $\uparrow$ & NC$\uparrow$ & HD  $\downarrow$ &  ECD $\downarrow$ & EF1 $\uparrow$ &$G_{avg}$ $\uparrow$ \\
    \midrule
    Dot prod & 5,982 & 5,982& \textbf{0.709} & \textbf{0.978} & \textbf{0.993} & {0.725}& \textbf{0.066} & \textbf{0.897} &\textbf{0.745}\\
    L2  & 5,982 & 39,543 & 7.97$\times 10^6$ & 0.951 & 0.977 & 578.65 & 0.267& 0.765 & 0.625\\
    Cos & 5,982 & 6,143 & 0.725 & 0.977 & 0.991 & \textbf{0.703} & 0.092 & 0.813 & 0.654 \\
    w/o aug & 5,982 & 6,143 & 0.738 & 0.976 & 0.991 & \textbf{0.703} & 0.077 & 0.862 & 0.656 \\
    \bottomrule
    \end{tabular}%
}\vspace{-5mm}
\end{center}
\end{table}

%% file: section/appendix.tex
\appendix
\title{NASM: Neural Anisotropic Surface Meshing: \\
Supplemental Document}
\settopmatter{printccs=false}
\maketitlesup

\section{Data Generation for Embedding and Meshing}
\label{data_gen}
\hspace{3mm}\textit{High-D Euclidean Embedding Computation.} For an arbitrary metric field $\mathbf{M}$ defined on a 3D surface domain $\Omega \subset \mathbb{R}^3$, there exists a high-d Euclidean embedded surface $\overline{\Omega} \subset \mathbb{R}^{\overline{m}}$~\cite{nash1954c1,Embedding2018}, where the mapping $\Omega \rightarrow \overline{\Omega}$ can be considered as a high-d transformation. Our computation process depends on the faithful high-d coordinates of the training meshes. To obtain that, we first follow the core idea of $\text{SIFHDE}^2$~\cite{Embedding2018} for the surface mesh case, which constructs the corresponding simplices in the target high-d space by deforming the tangent basis of each triangle in the mesh. The main advantages of this high-d embedding method are that we can automatically preserve the original 3D shape geometric features (such as sharp feature edges and corners) as well as avoid embedded self-intersections by using the strategy of keeping the original 3D coordinates, and only embedding additionally higher dimensions, i.e., ``anisotropic metric is traded as additional dimensions''.

For a triangle $\mathcal{F}\in \mathbb{R}^3$, let $\{\mathbf{v}_{i},\mathbf{v}_{j},\mathbf{v}_{k}\}$ denote its vertices. The basis of its tangent space $\mathbf{W}_{\mathcal{F}}$ can be given by its edge vectors, $\mathbf{W}_{\mathcal{F}} = [\mathbf{v}_{j}-\mathbf{v}_{i},\mathbf{v}_{k}-\mathbf{v}_{i}]$. The corresponding simplex in high-d space is $\mathcal{F}\in \mathbb{R}^{\overline{m}}$ where $\overline{m}\ge 3$ (as suggested in~\cite{Embedding2018}, $\overline{m} = 8$), and the basis of tangent space for $\mathcal{F}$ can be denoted by $\overline{\mathbf{W}}_{\mathcal{F}}=[\overline{\mathbf{v}}_{j}-\overline{\mathbf{v}}_{i},\overline{\mathbf{v}}_{k}-\overline{\mathbf{v}}_{i}]$. Their relation can be represented as:
\begin{align}
    \overline{\mathbf{W}}_{\mathcal{F}} = \mathbf{J}_{\mathcal{F}} \mathbf{W}_{\mathcal{F}},
\end{align}
where $\mathbf{J}_{\mathcal{F}}$ is the Jacobian transformation matrix of triangle $\mathcal{F}$, and $\mathbf{J}_{\mathcal{F}}^t \mathbf{J}_{\mathcal{F}} = \mathbf{M}_{\mathcal{F}}$.

\begin{figure}
    \centering
    \includegraphics[height=3.5cm]{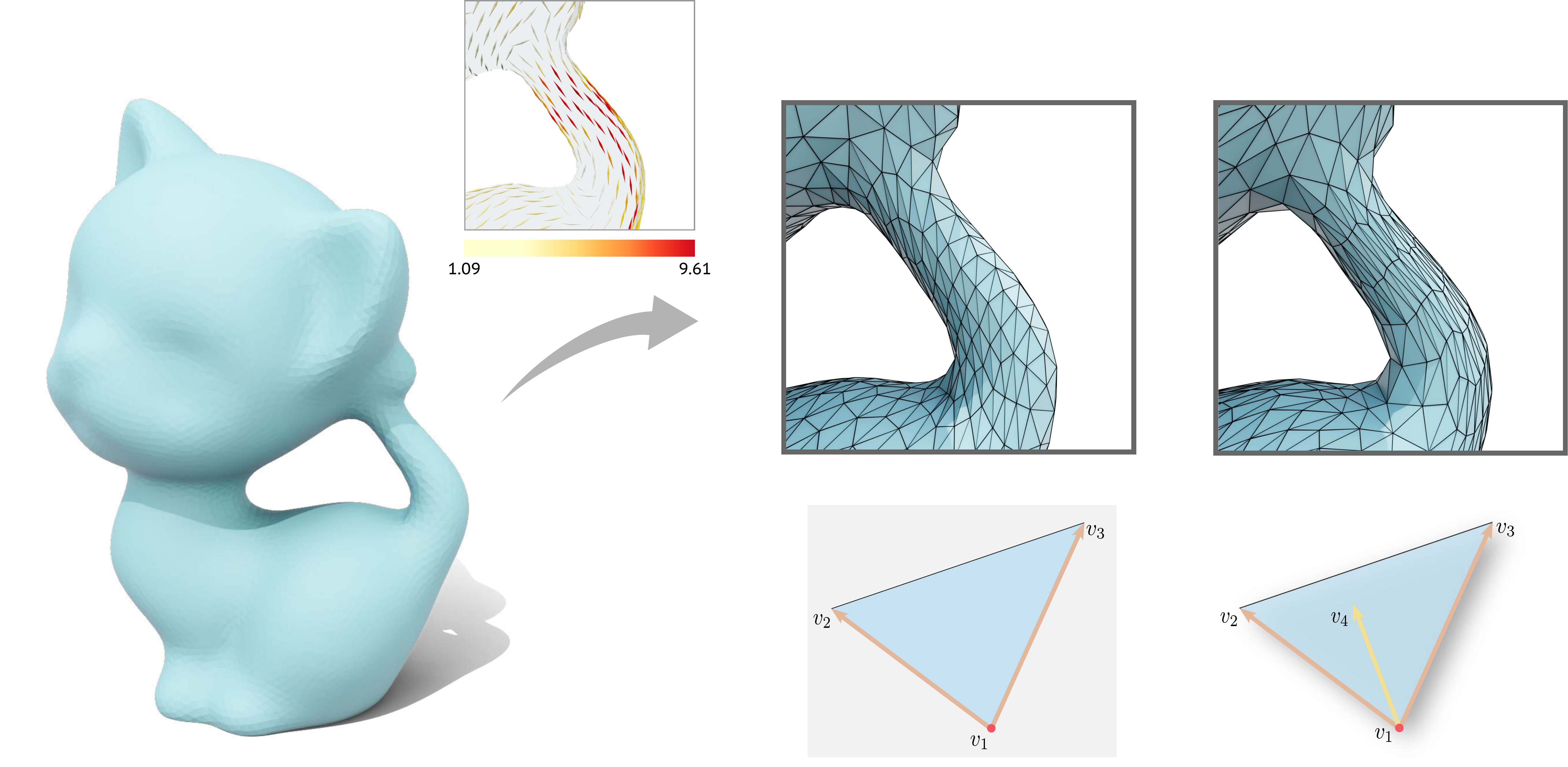}
    \caption{Illustration of adding a fourth vertex on the direction perpendicular to a triangle.}\vspace{-30mm} 
    \label{fig:perpen_triangle}
\end{figure}

However, this process could produce incorrect anisotropic direction on some part of the mesh surface, which makes the whole mesh object \textit{unstable} for training. The main reason is because $\mathbf{W}_{\mathcal{F}}$ is not a square matrix ($3\times 2$ as formed above), which cannot fully determine the deformation since the direction perpendicular to the triangle is not established as shown in Fig.~\ref{fig:perpen_triangle}.
To address this issue, inspired by~\cite{sumner2004deformation}, we construct a fourth vertex hypothetically on each triangle $\mathcal{F}$ in the direction perpendicular to the triangle as $\mathbf{v}_{l} = \frac{(\mathbf{v}_j-\mathbf{v}_i)\times (\mathbf{v}_k-\mathbf{v}_i)}{\sqrt{||(\mathbf{v}_j-\mathbf{v}_i)\times (\mathbf{v}_k-\mathbf{v}_i)||}}$, and result in a new basis of tangent plane $\mathbf{W}_{\mathcal{F}}'=[\mathbf{v}_{j}-\mathbf{v}_{i},\mathbf{v}_{k}-\mathbf{v}_{i},\mathbf{v}_{l}-\mathbf{v}_{i}]$. As for the high-d corresponding basis, we can discard the new vertex from the deformation result and keep the same notation $\overline{\mathbf{W}}_{\mathcal{F}}$ as above, since we only consider the surface case. Through the above computations, the ground truth data for training our neural high-d Euclidean embedding is generated.

\textit{Anisotropic Surface Mesh Generation.} After obtaining the high-d embedded coordinates $\overline{\mathbf{W}}_{\mathcal{F}}$, we can efficiently compute high-d isotropic restricted Voronoi diagrams (RVD) of object $\overline{\Omega}$ using the particle optimization technique~\cite{Embedding2018} or centroidal Voronoi tessellation (CVT) method~\cite{levy2013variational} in high-d space. Finally, the generated RVD and its dual restricted Delaunay triangulation can be mapped from the high-d embedding space to the original 3D space, where RVD and dual mesh can exhibit the desired anisotropy. In practice, we only need to get rid of the additional dimensions to get the back-mapped results since the embedding computation is based on the strategy of keeping the original 3D coordinates. Now, we can conduct the comparison experiments between the improved version of $\text{SIFHDE}^2$ and our NASM w.r.t. the surface accuracy and anisotropic mesh quality measurement.

\section{Combinatorial Structure of High-D Normal Metric CVT}
\label{comb1}
We consider a domain $\overline{\Omega}\in \mathbb{R}^{\overline{m}}$, where $\overline{\Omega}$ is a surface embedded in $\mathbb{R}^{\overline{m}}$. The combinatorial structure of $E_{hd}$ is determined by the high-d restricted Voronoi diagram (RVD), i.e., the intersection between the high-d Voronoi cells and the high-d embedded surface $\mathcal{\overline{S}}$. The RVD is first computed by an exact algorithm based on~\cite{yan2009RVD}. Then, each restricted Voronoi cell $\overline{\Omega}_{\mathbf{x}_0}\cap\mathcal{\overline{S}}$ is decomposed into a set of triangles called facets. The vertices of these facet triangles can have three different configurations and need to be treated differently in order to have the correct gradient when doing reverse-mode differentiation as follows:
\begin{itemize}
    \item $\mathbf{C}_1$: $\mathbf{\overline{V}}$ is a vertex of the original high-d embedded surface $\mathcal{\overline{S}}$;
    \item $\mathbf{C}_2$: $\mathbf{\overline{V}}$ is the intersection between an edge of the original high-d embedded surface $\mathcal{\overline{S}}$ and a bisector of two Voronoi cell sites $\mathbf{\overline{x}}_i$, $\mathbf{\overline{x}}_j$;
    \item $\mathbf{C}_3$: $\mathbf{\overline{V}}$ is the intersection between a triangle of the original high-d embedded surface $\mathcal{\overline{S}}$ and two sets of bisectors $\mathbf{\overline{x}}_i$, $\mathbf{\overline{x}}_j$ and $\mathbf{\overline{x}}_i$, $\mathbf{\overline{x}}_k$.
\end{itemize}

\section{Algebraic Structure of High-D Normal Metric CVT}
\label{algebra1}
The objective function of feature-sensitive high-d CVT is defined by adding a high-d normal metric $\overline{\mathbf{M}}^{\mathcal{T}}$ into the CVT energy in high-d:
\begin{align}\label{eq:hdcvt_energy}
    E_{hd}(\mathbf{\overline{X}}) &= \sum_i \underset{\overline{\Omega}_i\cap\mathcal{S}}{\int} \Big\|\overline{\mathbf{M}}^{\mathcal{T}}[\mathbf{\overline{y}}-\mathbf{\overline{x}}_i]\Big\|^2_2 d\mathbf{\overline{y}},\\
    &=\frac{|\mathcal{T}|}{
    \begin{pmatrix}
        n+p \\
        n 
    \end{pmatrix}} \sum_{\alpha+\beta+\gamma=p} \overline{\mathbf{U}_1^\alpha*\mathbf{U}_2^\beta*\mathbf{U}_3^\gamma},
\end{align}
\begin{align*}
    \text{where}:
    \begin{cases}
        \mathbf{U}_i &= \mathbf{\overline{M}}^{\mathcal{T}}(\mathbf{{C}}_i-\mathbf{\overline{x}}_0)\\
        \mathbf{\overline{V}}_1*\mathbf{\overline{V}}_2 &= [\overline{x}_1\overline{x}_2,\overline{y}_1\overline{y}_2,\overline{z}_1\overline{z}_2]^t\\
        \mathbf{\overline{V}}^\alpha &= \mathbf{\overline{V}}*\mathbf{\overline{V}}*\dots*\mathbf{\overline{V}} \;(\alpha \; \text{times})\\
        \overline{\mathbf{V}} &= \overline{x}+\overline{y}+\overline{z}
    \end{cases}.
\end{align*}
For RVD on the high-d embedded surface mesh, Voronoi cells are discretized to a set of facet triangles, denoted as $\mathcal{T}(\mathbf{\overline{x}}_0,\mathbf{{C}}_1,\mathbf{{C}}_2,\mathbf{{C}}_3)$. So the $\overline{\Omega}$ is the sum of the area of all the facets. A high-d CVT is a stable critical point of $E_{hd}$ during the optimization. 

As stated in~\cite{levy2010LpCVT}, given the gradient of a standard CVT energy $E$, $\nabla E = 2m_i(\mathbf{x}_i-\mathbf{g}_i)$, one cannot simply replace the centroid $\mathbf{g}_i$ and mass ${m}_i$ with their anisotropic counterparts to get the gradient of high-d CVT energy $E_{hd}$, since the anisotropy varies between two adjacent cells. The closed form for gradient of high-d CVT with normal metric over an integration simplex $\mathcal{T}$ is:
 \begin{equation}
 \label{eq:discretize1}
     \frac{dE^{\mathcal{T}}_{hd}(\mathbf{\overline{x}}_0,\mathbf{{C}}_1,\mathbf{{C}}_2,\mathbf{{C}}_3)}{d\mathbf{\overline{X}}} = 
     \frac{dE^{\mathcal{T}}_{hd}}{d\mathbf{\overline{x}}_0}+ \sum_{i=1,2,3} \frac{dE^{\mathcal{T}}_{hd}}{d\mathbf{{C}}_i}\frac{d\mathbf{{C}}_i}{d\mathbf{\overline{X}}},
 \end{equation}
 \begin{equation}
     \text{where}: \frac{dE^{\mathcal{T}}_{hd}}{d\mathbf{\overline{x}}_0} = -\frac{dE^{\mathcal{T}}_{hd}}{d\mathbf{{C}}_1}-\frac{dE^{\mathcal{T}}_{hd}}{d\mathbf{{C}}_2}-\frac{dE^{\mathcal{T}}_{hd}}{d\mathbf{{C}}_3},
 \end{equation}
where $\mathbf{\overline{X}}$ is the site point of the Voronoi cell. $\mathcal{T}(\mathbf{{C}}_1,\mathbf{{C}}_2,\mathbf{{C}}_3)$ is one of the facet triangles that compose the restricted Voronoi cell in the high-d embedding space.

\section{Auto Differentiation for High-D Normal Metric CVT}
Automatic differentiation is a computational technique that leverages the chain rule of calculus. Instead of computing gradients from explicit formula, automatic differentiation computes gradient starting from the function value, and tracing backwards along how the function value has been computed; and leverages the chain rule to compute the gradient. 

From Equation~(\ref{eq:discretize1}), the calculation of gradient can be separated to two parts: $\frac{dE^{\mathcal{T}}_{hd}}{d\mathbf{C}_i}$ and $\frac{d\mathbf{C}_i}{d\mathbf{\overline{X}}}$, and the total gradient can be assembled from these two expressions. In this subsection, we introduce how to calculate the $E^{\mathcal{T}}_{hd}$ and $\mathbf{C}_i$ in the forward pass in order to get the correct gradient $\frac{dE^{\mathcal{T}}_{hd}}{d\mathbf{C}_i}$ and $\frac{d\mathbf{C}_i}{d\mathbf{\overline{X}}}$ in the reverse pass.

For $E^{\mathcal{T}}_{hd}$, it can be discretized onto triangle facets that compose the restricted Voronoi cell, and the expression for this discretization can be shortly expressed as:
\begin{equation}
    E^{\mathcal{T}}_{hd}=|{\mathcal{T}}|F^{\mathcal{T}}_{hd},
\end{equation}
\begin{equation*}
    \text{where}: F^{\mathcal{T}}_{hd} = \sum_{\alpha+\beta+\gamma=p} \overline{\mathbf{U}_1^\alpha*\mathbf{U}_2^\beta*\mathbf{U}_3^\gamma},
\end{equation*}
where $|{\mathcal{T}}|$ denotes the area of current triangle facet, see Appendix A in~\cite{levy2010LpCVT} for details. 

\subsection{Derivative of $E^{\mathcal{T}}_{hd}$}
\label{a:derivative_e}
In \cite{levy2010LpCVT}, they used $|{\mathcal{T}}| = \frac{1}{2}\|N\|$ to calculate the area of $|{\mathcal{T}}|$, where $\|N\|$ is the length of cross product between two edges of $|{\mathcal{T}}|$. It is fine for 3D case, and they also derive the explicit gradient expression for $d|{\mathcal{T}}|$. However, this way of calculating triangle area cannot be extended to dimension higher than 3. As we targeting on higher dimension (e.g., $d$ = 8), we leverage Heron's formula for the area of a triangle in $\mathbb{R}^d$:
\begin{equation}\label{eq:heron}
    |{\mathcal{T}}| = \sqrt{s(s-a)(s-b)(s-c)},
\end{equation}
where $s = a+b+c$ and $a$, $b$, $c$ denote the length of three edges under the normal metric $\mathbf{\overline{M}}^{\mathcal{T}}$.

It is noted that our computation is more complicated and challenging than~\cite{levy2013variational}, which also uses Heron's formula to calculate the area of triangle in 6D, but without having the metric in their CVT energy function. So, their 6D CVT optimization does not require to compute the gradient of Heron's formula. Conversely, for our high-d normal metric CVT, $|{\mathcal{T}}|$ becomes dependent regards to $E^T_{hd}$:
\begin{equation}\label{eq:dfdc}
    \frac{dE^{\mathcal{T}}_{hd}}{d\mathbf{C}_i} = (\frac{dF^{\mathcal{T}}_{hd}}{d\mathbf{U}_i}|{\mathcal{T}}|+\frac{d|{\mathcal{T}}|}{d\mathbf{U}_i}F^{\mathcal{T}}_{hd})\mathbf{\overline{M}}^{\mathcal{T}},
\end{equation}
where $\mathbf{U}_i$ is defined in Equation~(\ref{eq:hdcvt_energy}). 

To use automatic differentiation on $\frac{dE^{\mathcal{T}}_{hd}}{d\mathbf{C}_i}$, we first set $\mathbf{U}_1$, $\mathbf{U}_2$, $\mathbf{U}_3$ as dependent variables, and $F^{\mathcal{T}}_{hd}$ can be calculated from Equation~(\ref{eq:hdcvt_energy}). Then $\frac{dF^{\mathcal{T}}_{hd}}{d\mathbf{U}_i}$ can be derived by backpropagation. For automatic differentiation on $\frac{d|{\mathcal{T}}|}{d\mathbf{U}_i}$, we can reuse the $\mathbf{U}_1$, $\mathbf{U}_2$, $\mathbf{U}_3$ to get $\mathbf{a} = \|\mathbf{U}_1-\mathbf{U}_2\|$, $\mathbf{b}=\|\mathbf{U}_2-\mathbf{U}_3\|$, $\mathbf{c}=\|\mathbf{U}_3-\mathbf{U}_1\|$ and use Equation~(\ref{eq:heron}) to calculate $|{\mathcal{T}}|$. Then we perform backpropagation to get $\frac{d|{\mathcal{T}}|}{d\mathbf{U}_i}$. Finally, $\frac{dE^{\mathcal{T}}_{hd}}{d\mathbf{C}_i}$ can be obtained by assembling all the values together following Equation~(\ref{eq:dfdc}).

The reason why we do not start the backpropagation from $E^{\mathcal{T}}_{hd}=|{\mathcal{T}}|F^{\mathcal{T}}_{hd}$ to derive $\frac{dE^{\mathcal{T}}_{hd}}{d\mathbf{C}_i}$ only, instead we take one step backward, is because the gradients for each $\mathbf{C}_i$ between these two ways could have some differences. So, our auto differentiation for the backpropagation makes the optimization convergence faster and more accurate. The detailed computational procedure for the derivative of $\mathbf{C}_i$ is provided in the following subsection.

\subsection{Derivative of $\mathbf{C}_i$} 
\label{a:derivative_c_i}
The partial derivative of $\frac{dE^{\mathcal{T}}_{hd}(\mathbf{\overline{x}}_0,\mathbf{C}_1,\mathbf{C}_2,\mathbf{C}_3)}{d\mathbf{\overline{X}}}$ also requires the partial gradient from vertices of facet triangles. For 3D case, every facet triangle $f$ can be described by their support plane $(\mathbf{N}_f,\mathbf{b}_f)$ of an equation $\mathbf{N}_f \mathbf{x}+\mathbf{b}_f=0$. The vertices of the facet triangles can be expressed by the intersections of bisectors and support planes according to its configuration type (i.e., $\mathbf{C}_1, \mathbf{C}_2, \mathbf{C}_3$) and $d\mathbf{C}_i$ can be obtained by taking derivative with respect to a ternary linear system. However, for a facet triangle embedded in the high-d space, its normal vector $\mathbf{N}_f\in \mathbb{R}^d$ cannot be easily obtained. We use Sutherland-Hodgman's re-entrant clipping~\cite{reentrant1974}. We first recall the re-entrant clipping algorithm. Given the bisector $[\mathbf{x}_0, \mathbf{x}_1]$ and an edge with two endpoints $\mathbf{A}$ and $\mathbf{B}$ that intersects the bisector, and the intersection $\mathbf{I}$ can be obtained by:
\begin{align}
    \mathbf{I} = \lambda_1 \mathbf{A}+\lambda_2 \mathbf{B},
\end{align}
\begin{align*}
    \text{where}: 
    \begin{cases}
        \lambda_1 = \frac{l_1}{|\mathbf{A}\mathbf{B}|} \\
        \lambda_2 = \frac{l_2}{|\mathbf{A}\mathbf{B}|} 
    \end{cases},
\end{align*}
where $l_1$ is the perpendicular distance from $\mathbf{A}$ to bisector $[\mathbf{\overline{x}}_0,\mathbf{\overline{x}}_1]$, $l_2$ is the perpendicular distance from $\mathbf{B}$ to bisector $[\mathbf{\overline{x}}_0,\mathbf{\overline{x}}_1]$. To calculate the perpendicular distance, we can calculate the distance between two parallel planes. The bisector can be defined as $d=-\frac{1}{2}(\mathbf{\overline{x}}_0+\mathbf{\overline{x}}_1)\cdot \overrightarrow{\mathbf{n}}$, and $\overrightarrow{\mathbf{n}}$ is the direction of $\overrightarrow{\mathbf{\overline{x}}_0\mathbf{\overline{x}}_1}$. The plane that passes through endpoint $\mathbf{A}$ and is parallel to the bisector is $d_1=-{\mathbf{A}}\cdot \overrightarrow{\mathbf{n}}$. $l_1$ and $l_2$ can be obtained by:
\begin{align}
    l_i = |d_i-d|.
\end{align}
For three types of configurations for $\mathbf{C}_i$, where $i = 1,2,3$, we should use the re-entrant clipping in different ways to ensure the gradient achieved from backpropagation is correct. For configuration $\mathbf{C}_1$, i.e., the original vertex of the surface $S$, it yields no derivative for site points $\mathbf{\overline{x}}_i$. For configuration $\mathbf{C}_2$, it can be seen as an original edge from the surface $S$ intersects with a bisector $[\mathbf{\overline{x}}_0,\mathbf{\overline{x}}_1]$, which is simply one pass of re-entrant clipping. We set $\mathbf{\overline{x}}_0$ and $\mathbf{\overline{x}}_1$ as dependent variables, two endpoints of the edges $e_1$ and $e_2$ as independent variables to get the intersection $\mathbf{C}_2$. Then, we can apply backpropagation from $\mathbf{C}_2$, and have the gradients of $\frac{d\mathbf{C}_2}{d\mathbf{\overline{x}}_0}$ and $\frac{d\mathbf{C}_2}{d\mathbf{\overline{x}}_1}$.

For configuration $\mathbf{C}_3$, the edges to be clipped with are not the original edges from surface $S$. The facet triangle is clipped by two bisectors $\mathbf{b}_1 =[\mathbf{\overline{x}}_0,\mathbf{\overline{x}}_1]$ and $\mathbf{b}_2=[\mathbf{\overline{x}}_0,\mathbf{\overline{x}}_2]$. Each bisector intersects with a facet triangle at its two edges, and the intersections are labeled as $\mathbf{I}_1^{b_1}$, $\mathbf{I}_2^{b_1}$ and $\mathbf{I}_1^{b_2}$, $\mathbf{I}_2^{b_2}$. The intersection between $[\mathbf{I}_1^{b_1},\mathbf{I}_2^{b_1}]$ and $[\mathbf{I}_1^{b_2},\mathbf{I}_2^{b_2}]$ coincides with $\mathbf{C}_3$. We can treat the process as re-entrant clipping that the bisector $\mathbf{b}_1=[\mathbf{\overline{x}}_0,\mathbf{\overline{x}}_1]$ intersects with edge $[\mathbf{I}_1^{b_2}, \mathbf{I}_2^{b_2}]$, and the gradient of $(\frac{d\mathbf{C}_3}{d\mathbf{\overline{x}}_0})^{b_1}$ and $\frac{d\mathbf{C}_3}{d\mathbf{\overline{x}}_1}$ can be obtained by backpropagation starting from $\mathbf{C}_3$. The same process should be applied to bisector $\mathbf{b}_2=[\mathbf{\overline{x}}_0,\mathbf{\overline{x}}_2]$ and edge $[\mathbf{I}_1^{b_1},\mathbf{I}_2^{b_1}]$, and we can obtain the gradient of $(\frac{d\mathbf{C}_3}{d\mathbf{\overline{x}}_0})^{b_2}$ and $\frac{d\mathbf{C}_3}{d\mathbf{\overline{x}}_2}$. The final gradients for $\mathbf{\overline{x}}_0$, $\mathbf{\overline{x}}_1$, and $\mathbf{\overline{x}}_2$ are $(\frac{d\mathbf{C}_3}{d\mathbf{\overline{x}}_0})^{b_1}+(\frac{d\mathbf{C}_3}{d\mathbf{\overline{x}}_0})^{b_2}$, $\frac{d\mathbf{C}_3}{d\mathbf{\overline{x}}_1}$, and $\frac{d\mathbf{C}_3}{d\mathbf{\overline{x}}_2}$, respectively.

\section{Additional Results}
\subsection{Thingi10k Dataset}
Tab.~\ref{tab:testset_add} shows a full version of quantitative comparison with our NASM method, NASM without high-d normal metric CVT, and $\text{SIFHDE}^2$ method~\cite{Embedding2018}. We use $100\%$, $80\%$, $60\%$, $40\%$, $20\%$, and $10\%$ of vertex count from input meshes and make the quantitative report. Our full NASM framework outperforms other cases / methods at all levels of the resolution. One of the main advantages of NASM (w/ or w/o NM CVT) over $\text{SIFHDE}^2$ method is: our results indicate that as long as most of the metrics in the dataset are accurate, NASM performs well during the inference. In contrast, $\text{SIFHDE}^2$ method is highly dependent on the quality of the metric for each individual model, which is more sensitive to inaccuracies of metrics. It is noted that the neural network can find a more general, accurate, and robust embedding than $\text{SIFHDE}^2$.

We also observe an increase of anisotropic mesh quality when using CVT for meshing, instead of using high-d normal metric CVT. Based on our analysis, it is actually a trade-off between anisotropic mesh quality and surface accuracy (feature preserving). Normal metric CVT can push the vertices towards sharp feature, which may distort the anisotropic metric direction and stretching ratio. So that the mesh quality evaluation may become lower. Fig.~\ref{fig:comparison_NASM} shows the qualitative comparison on NASM method, NASM without high-d normal metric CVT (with CVT), and $\text{SIFHDE}^2$ method~\cite{Embedding2018}. Meanwhile, the neural high-d embedding computation is much faster (about 1,500$\times$ speedup) than traditional $\text{SIFHDE}^2$ method. The normal metric CVT computation takes slightly more time than the general CVT computation.

\begin{figure*}
\centering
    \includegraphics[width=0.9\textwidth]{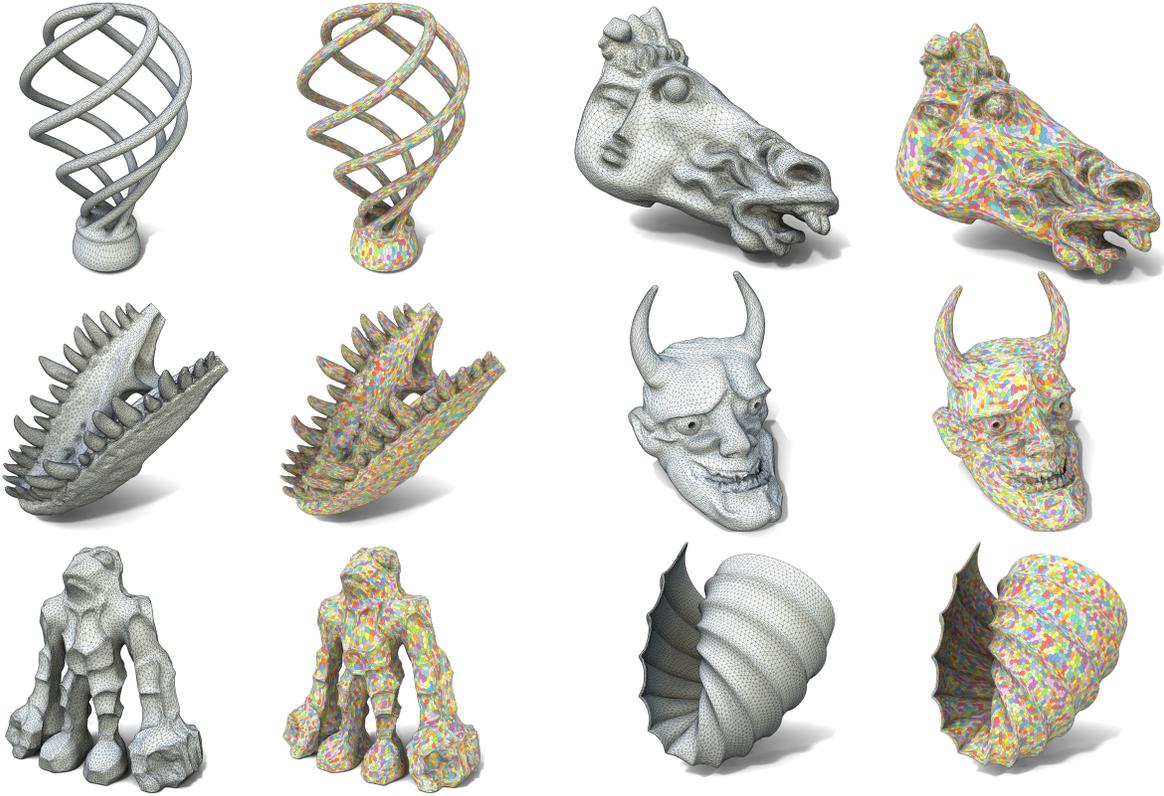}\vspace{-5pt}
    \caption{Anisotropic triangular meshes and their corresponding anisotropic RVD results of our NASM on some complicated surfaces from Thingi10k dataset.}
    \label{fig:RVD_add}
\end{figure*}

Fig.~\ref{fig:RVD_add} shows the additional anisotropic triangular meshes and their corresponding anisotropic RVD results of our NASM on some complicated surfaces from the Thingi10k dataset. The restricted Voronoi cells can well capture the sharp features and curvature anisotropies.

\subsection{MGN Dataset}
\begin{figure*}
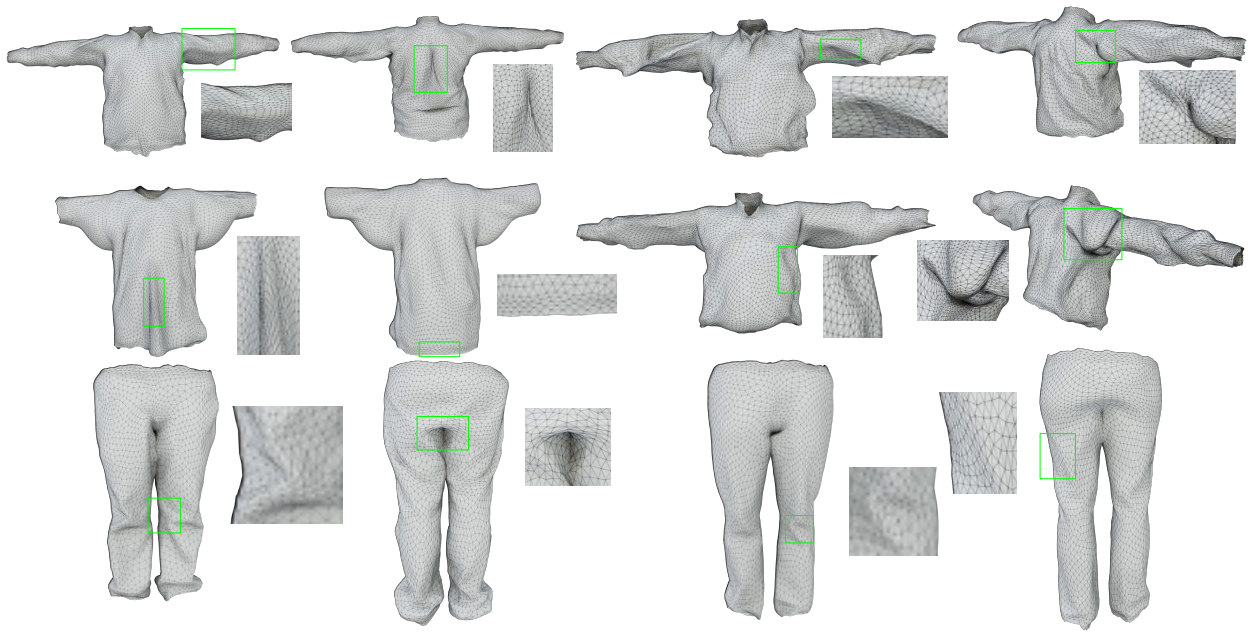

\centering
    \includegraphics[width=0.92\textwidth]{fig/mgn_top.pdf}\vspace{-5pt}
    \includegraphics[width=0.80\textwidth]{fig/mgn_bot.pdf}\vspace{-5pt}
    \caption{Additional anisotropic surface meshing results of our NASM method on an unseen testing dataset, e.g., MGN dataset. Several long and short sleeve tops and pants meshes are shown to well capture the open boundaries and detailed cloth wrinkles and folds.}
    \label{fig:mgn}
\end{figure*}

Tab.~\ref{tab:comparison_MGN} shows the quantitative evaluation on Multi-Garment Net (MGN) dataset~\cite{bhatnagar2019mgn} with 154 cloth models (including 96 pants and 58 tops). Fig.~\ref{fig:mgn} shows the additional anisotropic surface meshing results of our NASM method on MGN dataset. The examples of our results can well capture the open boundaries, and detailed cloth wrinkles and folds from tops and pants models.

\subsection{FAUST Dataset}
\begin{figure*}
\centering
    \includegraphics[width=0.92\textwidth]{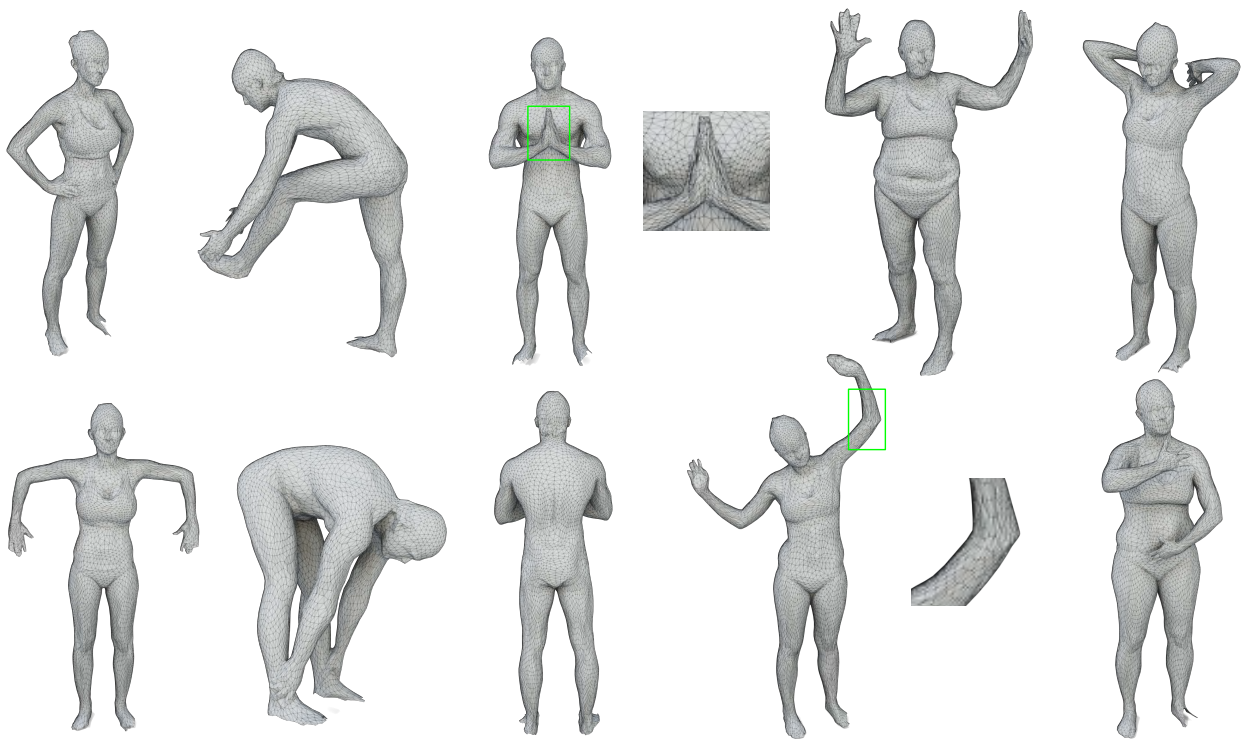}\vspace{-5pt}
    \caption{Additional anisotropic surface meshing results of our NASM method on another unseen testing dataset, e.g., FAUST dataset. Several human subject meshes in different poses are shown to well capture geometric anisotropies and features around hands, arms, legs, and thin clothes wrinkles on 3D human models.}
    \label{fig:faust}
\end{figure*}

Tab~\ref{tab:comparison_FAUST} shows the quantitative evaluation on FAUST dataset~\cite{faust2014} with 200 human scan models of 10 different subjects in 30 different poses. Each scan is a high-resolution and non-watertight triangular mesh. Even though there are some issues in the original meshes, such as feet are not complete, and some fingers are attached to each other, etc., our NASM method can still handle these non-manifold meshes. We use QEM~\cite{garland1997surface} to reduce the mesh resolution (e.g., $5\%$ of the original mesh elements) to feed into the neural network. The evaluation is conducted between NASM results and reduced-resolution meshes. Fig.~\ref{fig:faust} shows the anisotropic surface meshing results of our NASM method on FAUST dataset. Our results demonstrate promising results to capture geometric anisotropies and features around hands, arms, legs, and thin clothes wrinkles on 3D human models.

\subsection{Synthetic Models}
\begin{figure*}
\centering
    \includegraphics[width=0.8\textwidth]{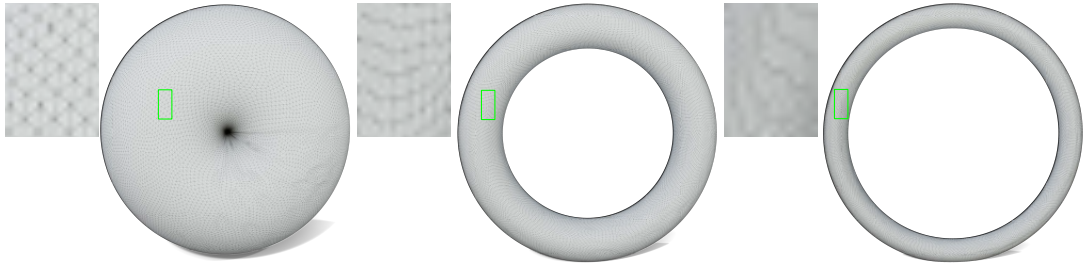}\vspace{-5pt}
    \caption{The qualitative results of our NASM method on synthetic torus models with different stretching ratios. From left to right, the stretching ratios are: 1:2, 1:6, and 1:11, respectively.}
    \label{fig:torus}
\end{figure*}

Fig.~\ref{fig:torus} shows the anisotropic surface meshing results of our NASM method on synthetic mathematical torus models with different stretching ratios, e.g., 1:2, 1:6, and 1:11. Our results show that the final mesh elements can well match the stretching ratios and directions of the curvature metric fields.

\section{Additional Analyses}
\subsection{Timing}
\begin{figure*}
    \centering
    \includegraphics[width=0.75\textwidth]{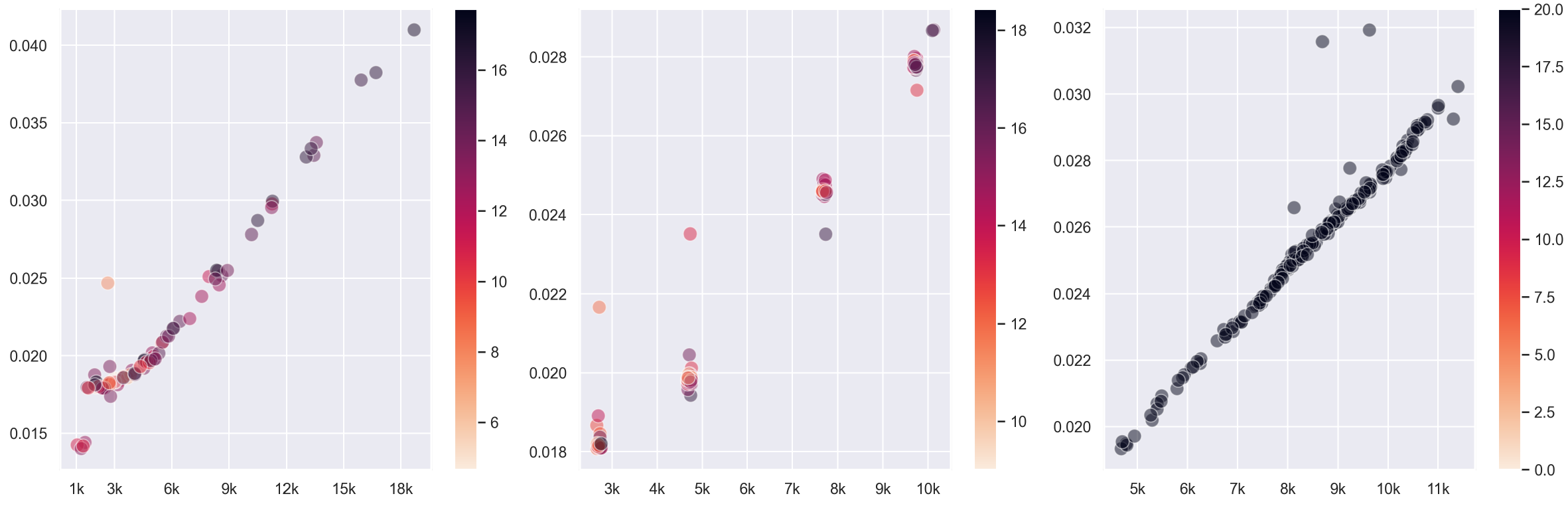}\vspace{-2mm}
    \caption{Inference times of NASM embedding computation with respect to the number of input mesh vertices from three datasets (left to right: 80 models from Thingi10k, 154 models from MGN, and 200 models from FAUST). The colormap indicates the maximum anisotropic stretching ratio of the corresponding mesh.}
    \label{fig:timing}
\end{figure*}

Fig.~\ref{fig:timing} shows the inference times of NASM embedding computation with respect to the input number of vertices for all three testing datasets, i.e., Thingi10k dataset (80 models), MGN dataset (154 models), and FAUST dataset (200 models). The colormap indicates the maximum anisotropic stretching ratio of the corresponding mesh. It is clear to see that the relationship between the inference time and the input number of mesh vertices is consistently linear among all three datasets.

\subsection{Ablation Study}
\begin{figure*}
\centering
    \includegraphics[width=0.75\textwidth]{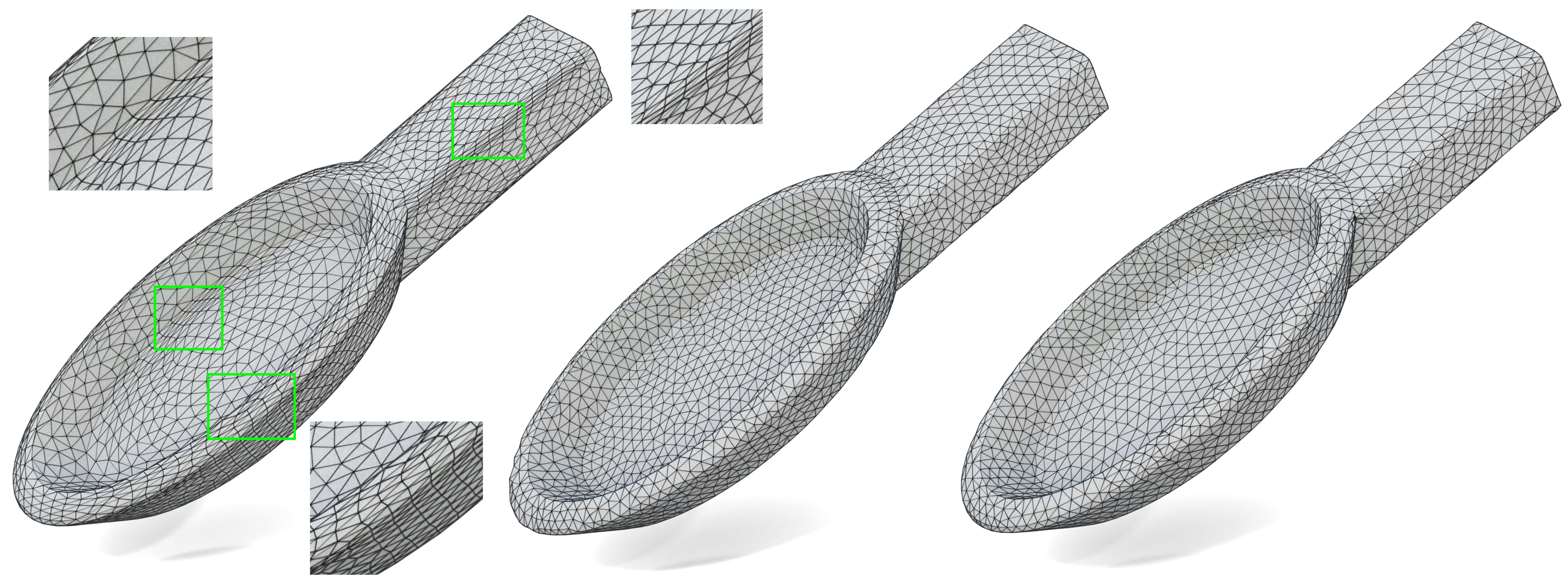}\vspace{-5pt}
    \caption{The qualitative results of ablation study on our NASM method with different weights for Laplacian loss: 0.1, 0.3, and 0.5 (from left to right).}
    \label{fig:ablation_wlap}
\end{figure*}

Tab.~\ref{tab:comparison_laplacian} shows the analysis of $w_{lap}$ value setting in high-d Euclidean embedding loss function, i.e., Equation (4) in the main paper. We compare with different weights of Laplacian loss, such as 0.1, 0.3, and 0.5 on Thingi10k dataset. Fig.~\ref{fig:ablation_wlap} shows the qualitative comparison with different weights of Laplacian loss $w_{lap}$. Finally, it is clear to see that 0.1 is the optimal setting for $w_{lap}$ in our task. 

\begin{figure*}
\centering
    \includegraphics[width=0.92\textwidth]{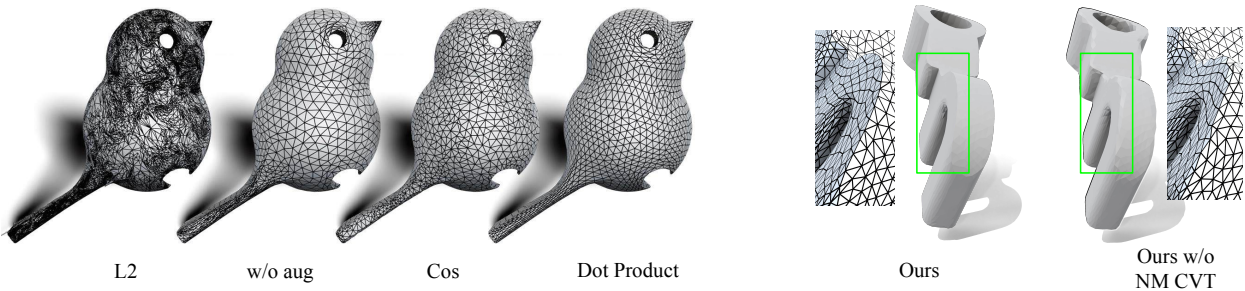}\vspace{-5pt}
    \caption{The qualitative results of ablation study on our NASM method with different loss functions, without data augmentation, and without high-d normal metric CVT.}
    \label{fig:ablation_loss}
\end{figure*}

Fig.~\ref{fig:ablation_loss} shows the qualitative results of ablation study on our NASM method with different loss functions, w/o data augmentation, and w/o high-d normal metric CVT. Our proposed dot product loss and full version of NASM demonstrate better performance than other cases.

\subsection{Failure Case}
\begin{figure*}
\centering
    \begin{overpic}[width=0.92\linewidth]{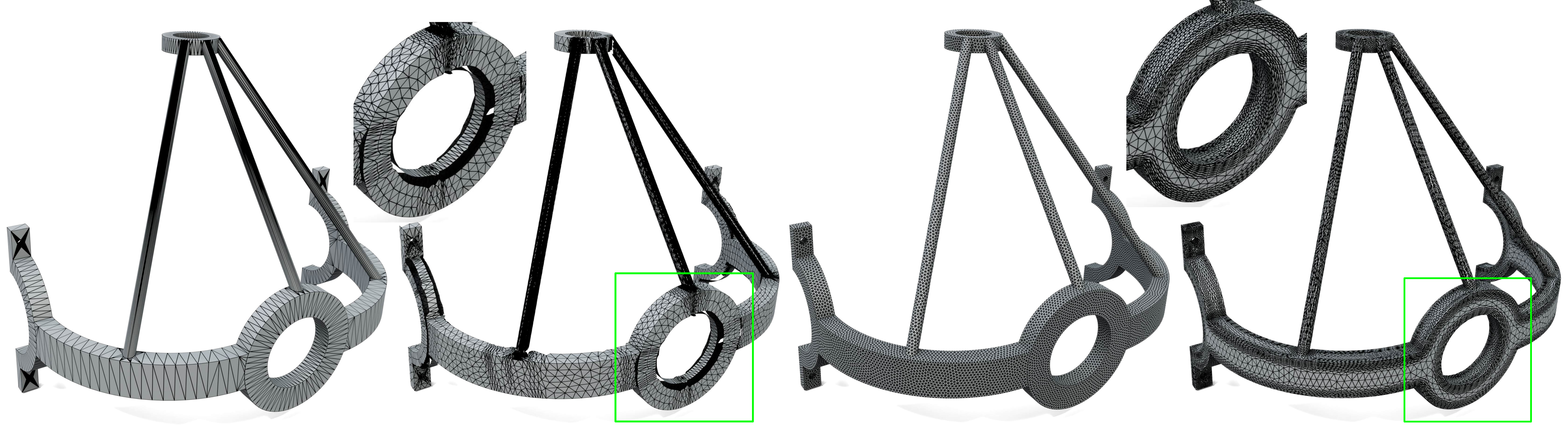}
    \put(22,0.1){{Sparse Input}}
    \put(75,0.1){{Dense Input}}
    \end{overpic}
    \caption{Failure case: our method cannot handle the CAD-like model with a triangulation with extremely sparse vertices, e.g., 2,406 vertices (left); while our method can successfully handle a denser triangulation, e.g., 20,000 vertices (right).}
    \label{fig:limitation}
\end{figure*}

Fig.~\ref{fig:limitation} shows that our NASM method may fail on a CAD-like model with very sparse input vertices, since a high-quality curvature-based embedding is very challenging to be predicted in such cases. Once we increase the mesh resolution to some extent, we can obtain a good-quality anisotropic mesh result.

\begin{table*}
\caption{Quantitative comparison with our NASM method, NASM without high-d normal metric CVT, and $\text{SIFHDE}^2$ method~\cite{Embedding2018} on 80 models selected from Thingi10k dataset, including smooth surfaces and surfaces with sharp and weak features. All the evaluate metrics are average values of all models from the dataset. The best results are highlighted in bold per different $\#V_{out}$. Note: $\#V_{in}$ is the average number of vertices of all input meshes, $\#V_{out}$ is the average number of vertices of all output meshes, `Stretch' is the average anisotropic stretching ratios of all models, CD ($\times 10^5$), HD ($\times 10^2$), ECD ($\times 10^2$).}\vspace{-2mm}
\label{tab:testset_add}
\begin{center}
\resizebox{0.76\textwidth}{!}{%
    \begin{tabular}{lllllllllllll}
    \toprule
     Method &  $\#V_{in}$ & $\#V_{out}$ & Stretch & CD $\downarrow$ & F1 $\uparrow$ & NC $\uparrow$ & HD $\downarrow$ &ECD $\downarrow$ & EF1 $\uparrow$ & $T_{em}$ (s) $\downarrow$ & $G_{avg}$ $\uparrow$ & $T_{mesh}$ (s) $\downarrow$ \\
    \midrule
    NASM & 5,982 & 5,982  & 12.736 & \textbf{0.709} & \textbf{0.978} & \textbf{0.993} & \textbf{0.725}  & \textbf{0.066} & \textbf{0.897} & \textbf{0.029} & 0.745 & 14.022 \\
    & 5,982 & 4,786 & 12.736& \textbf{0.712} & \textbf{0.978} & \textbf{0.992} & \textbf{0.739} & \textbf{0.074} & \textbf{0.882} & \textbf{0.029} & 0.747 & 11.648 \\
    & 5,982 & 3,590 & 12.736& \textbf{0.720} & \textbf{0.978} & \textbf{0.991} & \textbf{0.779} & \textbf{0.086} & \textbf{0.850} &\textbf{0.029} & 0.748 & 7.366 \\
    & 5,982 & 2,395  &12.736 & \textbf{0.741} & \textbf{0.976} & \textbf{0.990} & \textbf{0.842} & \textbf{0.102} & \textbf{0.804} & \textbf{0.029} & 0.749 & 5.751 \\
    & 5,982 & 1,202  & 12.736& \textbf{0.882} & \textbf{0.967} & \textbf{0.984} & \textbf{1.127} & \textbf{0.148} & \textbf{0.676} & \textbf{0.029} & 0.744 & 3.499 \\
    &5,982  & 608  & 12.736& \textbf{1.538} & \textbf{0.928} & \textbf{0.974} & \textbf{1.774} & \textbf{0.207} & \textbf{0.501} & \textbf{0.029} & 0.732 & \textbf{2.092} \\
    \midrule
    NASM  & 5,982 & 5,982 & 12.736 & 0.779 & 0.972 & 0.989 & 0.882  & 0.137 & 0.687 & \textbf{0.029} & \textbf{0.758} & 3.780 \\
    w/o NM CVT & 5,982 & 4,786 & 12.736& 0.812 & 0.968 & 0.988 & 0.940 & 0.145 & 0.635 & \textbf{0.029} & \textbf{0.760} & 3.587 \\
    (w/ CVT) &5,982  & 3,590  &12.736 & 0.875 & 0.963 & 0.987 & 1.055 & 0.155 & 0.571 & \textbf{0.029} & \textbf{0.764} & \textbf{3.354} \\  
    & 5,982 & 2,395 &12.736 & 1.026 & 0.951 & 0.983 & 1.207 & 0.173 & 0.469 & \textbf{0.029} & \textbf{0.768} & \textbf{3.185} \\
    & 5,982 & 1,202  & 12.736& 1.684 & 0.905 & 0.975 & 1.613 & 0.215 & 0.290 & \textbf{0.029} & \textbf{0.776} & 3.127 \\
    & 5,982 & 608 &12.736 & 3.692 & 0.779 & 0.962 & 2.352 & 0.267 & 0.170 & \textbf{0.029} & \textbf{0.779} & 2.922 \\
     \midrule
    $\text{SIFHDE}^2$ & 5,982 & 5,982 & 12.736 & 0.808 & 0.969 & 0.988 & 0.949 & 0.146 & 0.612 & 49.25 & 0.729 & \textbf{3.718} \\
    & 5,982 & 4,786 &12.736 & 0.850 & 0.965 & 0.987 & 1.008 & 0.154 & 0.561 & 49.25 & 0.731 & \textbf{3.525} \\
    & 5,982 & 3,590 & 12.736& 0.928 & 0.959 & 0.985 & 1.145 & 0.166 & 0.487 & 49.25 & 0.732 & 3.441 \\
    & 5,982 & 2,395 & 12.736& 1.109 & 0.945 & 0.982 & 1.330 & 0.185 & 0.388 & 49.25 & 0.733 & 3.252 \\
    & 5,982 & 1,202 &12.736 & 1.878 & 0.893 & 0.975 & 1.848 & 0.222 & 0.249 & 49.25 & 0.734 & \textbf{3.005} \\
    & 5,982 & 608 &12.736 & 4.144 & 0.766 & 0.963 & 2.674 & 0.270 & 0.157 & 49.25 & 0.730 & 2.935 \\
    \bottomrule
    \end{tabular}%
}
\end{center}
\end{table*}

\begin{figure*}
\centering
\begin{overpic}[width=0.88\textwidth]{fig/sifhde_more_1.pdf}
    \put(0.1,24.0){{NASM:}}
\end{overpic}\vspace{7pt}
\begin{overpic}[width=0.88\textwidth]{fig/sifhde_more_2.pdf}
    \put(0.1,25.0){{NASM w/o NM-CVT:}}
\end{overpic}
\begin{overpic}[width=0.88\textwidth]{fig/sifhde_more_3.pdf}\vspace{-8pt}
    \put(0.1,24.0){{SIFHDE$^2$:}}
\end{overpic}\vspace{-2mm}
    \caption{Qualitative comparison with NASM method, NASM without high-d normal metric CVT (with CVT), and $\text{SIFHDE}^2$ method~\cite{Embedding2018} (from top row to bottom row).}
    \label{fig:comparison_NASM}
\end{figure*}

\begin{table*}
\caption{Quantitative evaluation on Multi-Garment Net (MGN) dataset with 154 cloth models (including 96 pants and 58 tops). All the evaluate metrics are average values of all
models from the dataset. Note: CD ($\times 10^5$), HD ($\times 10^2$), ECD ($\times 10^2$).}
\label{tab:comparison_MGN}
\begin{center}
\resizebox{0.6\textwidth}{!}{%
    \begin{tabular}{llllllllll}
    \toprule
     \#Meshes &$\#V_{in}$ & $\#V_{out}$ &CD $\downarrow$& F1 $\uparrow$ & NC$\uparrow$ & HD  $\downarrow$  &ECD $\downarrow$ & EF1 $\uparrow$ &$G_{avg}$ $\uparrow$ \\
    \midrule
    154 & 6,033 & 6,020 & 0.549 & 0.993 & 0.993 & 1.216& 0.098 & 0.788 &0.725\\
    \bottomrule
    \end{tabular}%
}
\end{center}
\end{table*}

\begin{table*}
\caption{Quantitative evaluation on FAUST dataset with 200 human scans models of 10 different subjects in 30 different poses. All the evaluate metrics are average values of all
models from the dataset. Note: CD ($\times 10^5$), HD ($\times 10^2$), ECD ($\times 10^2$).}
\label{tab:comparison_FAUST}
\begin{center}
\resizebox{0.55\textwidth}{!}{%
    \begin{tabular}{lllllllll}
    \toprule
     \#Meshes &$\#V_{in}$ & $\#V_{out}$ &CD $\downarrow$& F1 $\uparrow$ & NC$\uparrow$ & HD  $\downarrow$  &ECD  $\downarrow$ & EF1 $\uparrow$ \\
    \midrule
    200 & 8,522 & 8,505 & 0.377 & 0.997 & 0.987 & 2.390& 0.099 & 0.791\\
    \bottomrule
    \end{tabular}%
}
\end{center}
\end{table*}

\begin{table*}
\caption{Ablation study on the weight of Laplacian loss with 0.1, 0.3, and 0.5 on 80 models selected from Thingi10k dataset. All the evaluate metrics are average values of all
models from the dataset. The best results are highlighted in bold. Note: CD ($\times 10^5$), HD ($\times 10^2$), ECD ($\times 10^2$).}
\label{tab:comparison_laplacian}
\begin{center}
\resizebox{0.6\textwidth}{!}{%
    \begin{tabular}{llllllllll}
    \toprule
    $w_{lap}$ &$\#V_{in}$ & $\#V_{out}$ &CD $\downarrow$& F1 $\uparrow$ & NC$\uparrow$ & HD  $\downarrow$ &ECD $\downarrow$ & EF1 $\uparrow$ &$G_{avg}$ $\uparrow$ \\
    \midrule
    0.1  & 5,982 & 5,982& \textbf{0.709} & \textbf{0.978} & \textbf{0.993} & \textbf{0.725}& \textbf{0.066} & \textbf{0.897} &\textbf{0.745}\\
    0.3  & 5,982 & 16,398 & 1.677 & 0.973 & 0.991 & 2.148  & 0.113 & 0.773 & 0.700\\
    0.5 & 5,982 & 35,057 & 2.146 & 0.967 & 0.986 & 6.195  & 0.200 & 0.518 & 0.534 \\
    \bottomrule
    \end{tabular}%
}
\end{center}
\end{table*}


%% file: sample-acmtog-SIGGRAPH-submission.bbl

\begin{thebibliography}{75}


\ifx \showCODEN    \undefined \def \showCODEN     #1{\unskip}     \fi
\ifx \showDOI      \undefined \def \showDOI       #1{#1}\fi
\ifx \showISBNx    \undefined \def \showISBNx     #1{\unskip}     \fi
\ifx \showISBNxiii \undefined \def \showISBNxiii  #1{\unskip}     \fi
\ifx \showISSN     \undefined \def \showISSN      #1{\unskip}     \fi
\ifx \showLCCN     \undefined \def \showLCCN      #1{\unskip}     \fi
\ifx \shownote     \undefined \def \shownote      #1{#1}          \fi
\ifx \showarticletitle \undefined \def \showarticletitle #1{#1}   \fi
\ifx \showURL      \undefined \def \showURL       {\relax}        \fi
\providecommand\bibfield[2]{#2}
\providecommand\bibinfo[2]{#2}
\providecommand\natexlab[1]{#1}
\providecommand\showeprint[2][]{arXiv:#2}

\bibitem[Alauzet and Loseille(2010)]%
        {Alauzet:JCP2010}
\bibfield{author}{\bibinfo{person}{Fr{\'e}d{\'e}ric Alauzet} {and} \bibinfo{person}{Adrien Loseille}.} \bibinfo{year}{2010}\natexlab{}.
\newblock \showarticletitle{High-Order Sonic Boom Modeling Based on Adaptive Methods}.
\newblock \bibinfo{journal}{\emph{J. Comput. Phys.}} \bibinfo{volume}{229}, \bibinfo{number}{3} (\bibinfo{year}{2010}), \bibinfo{pages}{561--593}.
\newblock


\bibitem[Alliez et~al\mbox{.}(2003)]%
        {Alliez2003}
\bibfield{author}{\bibinfo{person}{Pierre Alliez}, \bibinfo{person}{David Cohen-Steiner}, \bibinfo{person}{Olivier Devillers}, \bibinfo{person}{Bruno L\'{e}vy}, {and} \bibinfo{person}{Mathieu Desbrun}.} \bibinfo{year}{2003}\natexlab{}.
\newblock \showarticletitle{Anisotropic Polygonal Remeshing}.
\newblock \bibinfo{journal}{\emph{ACM Transactions on Graphics}} \bibinfo{volume}{22}, \bibinfo{number}{3} (\bibinfo{year}{2003}), \bibinfo{pages}{485–493}.
\newblock


\bibitem[Amestoy et~al\mbox{.}(2001)]%
        {amestoy2001fully}
\bibfield{author}{\bibinfo{person}{Patrick~R Amestoy}, \bibinfo{person}{Iain~S Duff}, \bibinfo{person}{Jean-Yves L'Excellent}, {and} \bibinfo{person}{Jacko Koster}.} \bibinfo{year}{2001}\natexlab{}.
\newblock \showarticletitle{A fully asynchronous multifrontal solver using distributed dynamic scheduling}.
\newblock \bibinfo{journal}{\emph{SIAM J. Matrix Anal. Appl.}} \bibinfo{volume}{23}, \bibinfo{number}{1} (\bibinfo{year}{2001}), \bibinfo{pages}{15--41}.
\newblock


\bibitem[Bhatnagar et~al\mbox{.}(2019)]%
        {bhatnagar2019mgn}
\bibfield{author}{\bibinfo{person}{Bharat~Lal Bhatnagar}, \bibinfo{person}{Garvita Tiwari}, \bibinfo{person}{Christian Theobalt}, {and} \bibinfo{person}{Gerard Pons-Moll}.} \bibinfo{year}{2019}\natexlab{}.
\newblock \showarticletitle{Multi-Garment Net: Learning to Dress 3D People from Images}. In \bibinfo{booktitle}{\emph{{IEEE} International Conference on Computer Vision}}.
\newblock


\bibitem[Bogo et~al\mbox{.}(2014)]%
        {bogo2014faust}
\bibfield{author}{\bibinfo{person}{Federica Bogo}, \bibinfo{person}{Javier Romero}, \bibinfo{person}{Matthew Loper}, {and} \bibinfo{person}{Michael~J Black}.} \bibinfo{year}{2014}\natexlab{}.
\newblock \showarticletitle{{FAUST}: Dataset and Evaluation for {3D} Mesh Registration}. In \bibinfo{booktitle}{\emph{Proceedings of the IEEE conference on computer vision and pattern recognition}}. \bibinfo{pages}{3794--3801}.
\newblock


\bibitem[Boissonnat et~al\mbox{.}(2015a)]%
        {boissonnat2015a}
\bibfield{author}{\bibinfo{person}{Jean-Daniel Boissonnat}, \bibinfo{person}{Kan-Le Shi}, \bibinfo{person}{Jane Tournois}, {and} \bibinfo{person}{Mariette Yvinec}.} \bibinfo{year}{2015}\natexlab{a}.
\newblock \showarticletitle{Anisotropic Delaunay Meshes of Surfaces}.
\newblock \bibinfo{journal}{\emph{ACM Transactions on Graphics}} \bibinfo{volume}{34}, \bibinfo{number}{2} (\bibinfo{year}{2015}).
\newblock


\bibitem[Boissonnat et~al\mbox{.}(2015b)]%
        {boissonnat2015b}
\bibfield{author}{\bibinfo{person}{Jean-Daniel Boissonnat}, \bibinfo{person}{Camille Wormser}, {and} \bibinfo{person}{Mariette Yvinec}.} \bibinfo{year}{2015}\natexlab{b}.
\newblock \showarticletitle{Anisotropic Delaunay Mesh Generation}.
\newblock \bibinfo{journal}{\emph{SIAM J. Comput.}} \bibinfo{volume}{44}, \bibinfo{number}{2} (\bibinfo{year}{2015}), \bibinfo{pages}{467--512}.
\newblock


\bibitem[Bossen and Heckbert(1996)]%
        {Bossen1996}
\bibfield{author}{\bibinfo{person}{Frank~J Bossen} {and} \bibinfo{person}{Paul~S Heckbert}.} \bibinfo{year}{1996}\natexlab{}.
\newblock \showarticletitle{A Pliant Method for Anisotropic Mesh Generation}.
\newblock   \bibinfo{volume}{63} (\bibinfo{year}{1996}), \bibinfo{pages}{76}.
\newblock


\bibitem[Bronstein et~al\mbox{.}(2017)]%
        {Bronstein_2017}
\bibfield{author}{\bibinfo{person}{Michael~M. Bronstein}, \bibinfo{person}{Joan Bruna}, \bibinfo{person}{Yann LeCun}, \bibinfo{person}{Arthur Szlam}, {and} \bibinfo{person}{Pierre Vandergheynst}.} \bibinfo{year}{2017}\natexlab{}.
\newblock \showarticletitle{Geometric Deep Learning: Going beyond Euclidean data}.
\newblock \bibinfo{journal}{\emph{IEEE Signal Processing Magazine}} \bibinfo{volume}{34}, \bibinfo{number}{4} (\bibinfo{date}{July} \bibinfo{year}{2017}), \bibinfo{pages}{18–42}.
\newblock


\bibitem[Bruna et~al\mbox{.}(2014)]%
        {Bruna2013}
\bibfield{author}{\bibinfo{person}{Joan Bruna}, \bibinfo{person}{Wojciech Zaremba}, \bibinfo{person}{Arthur Szlam}, {and} \bibinfo{person}{Yann Lecun}.} \bibinfo{year}{2014}\natexlab{}.
\newblock \showarticletitle{Spectral Networks and Locally Connected Networks on Graphs}.
\newblock  (\bibinfo{year}{2014}).
\newblock


\bibitem[Budninskiy et~al\mbox{.}(2016)]%
        {budninskiy2016optimal}
\bibfield{author}{\bibinfo{person}{Max Budninskiy}, \bibinfo{person}{Beibei Liu}, \bibinfo{person}{Fernando De~Goes}, \bibinfo{person}{Yiying Tong}, \bibinfo{person}{Pierre Alliez}, {and} \bibinfo{person}{Mathieu Desbrun}.} \bibinfo{year}{2016}\natexlab{}.
\newblock \showarticletitle{Optimal Voronoi tessellations with Hessian-based anisotropy}.
\newblock \bibinfo{journal}{\emph{ACM Transactions on Graphics}} \bibinfo{volume}{35}, \bibinfo{number}{6} (\bibinfo{year}{2016}), \bibinfo{pages}{1--12}.
\newblock


\bibitem[Carpenter et~al\mbox{.}(2015)]%
        {carpenter2015stan}
\bibfield{author}{\bibinfo{person}{Bob Carpenter}, \bibinfo{person}{Matthew~D. Hoffman}, \bibinfo{person}{Marcus Brubaker}, \bibinfo{person}{Daniel Lee}, \bibinfo{person}{Peter Li}, {and} \bibinfo{person}{Michael Betancourt}.} \bibinfo{year}{2015}\natexlab{}.
\newblock \bibinfo{title}{The Stan Math Library: Reverse-Mode Automatic Differentiation in C++}.
\newblock
\newblock
\showeprint[arxiv]{1509.07164}


\bibitem[Chen et~al\mbox{.}(2007)]%
        {chen2007optimal}
\bibfield{author}{\bibinfo{person}{Long Chen}, \bibinfo{person}{Pengtao Sun}, {and} \bibinfo{person}{Jinchao Xu}.} \bibinfo{year}{2007}\natexlab{}.
\newblock \showarticletitle{Optimal Anisotropic Meshes for Minimizing Interpolation Errors in Lp-norm}.
\newblock \bibinfo{journal}{\emph{Math. Comp.}} \bibinfo{volume}{76}, \bibinfo{number}{257} (\bibinfo{year}{2007}), \bibinfo{pages}{179--204}.
\newblock


\bibitem[Chen and Xu(2004)]%
        {chen2004optimal}
\bibfield{author}{\bibinfo{person}{Long Chen} {and} \bibinfo{person}{Jin-chao Xu}.} \bibinfo{year}{2004}\natexlab{}.
\newblock \showarticletitle{Optimal Delaunay Triangulations}.
\newblock \bibinfo{journal}{\emph{Journal of Computational Mathematics}} (\bibinfo{year}{2004}), \bibinfo{pages}{299--308}.
\newblock


\bibitem[Chen et~al\mbox{.}(2021)]%
        {Chen_2021}
\bibfield{author}{\bibinfo{person}{Zhiqin Chen}, \bibinfo{person}{Andrea Tagliasacchi}, {and} \bibinfo{person}{Hao Zhang}.} \bibinfo{year}{2021}\natexlab{}.
\newblock \showarticletitle{Learning Mesh Representations via Binary Space Partitioning Tree Networks}.
\newblock \bibinfo{journal}{\emph{IEEE Transactions on Pattern Analysis and Machine Intelligence}} (\bibinfo{year}{2021}), \bibinfo{pages}{1–1}.
\newblock


\bibitem[Dassi et~al\mbox{.}(2014)]%
        {DASSI2014}
\bibfield{author}{\bibinfo{person}{Franco Dassi}, \bibinfo{person}{Andrea Mola}, {and} \bibinfo{person}{Hang Si}.} \bibinfo{year}{2014}\natexlab{}.
\newblock \showarticletitle{Curvature-Adapted Remeshing of CAD Surfaces}.
\newblock \bibinfo{journal}{\emph{Procedia Engineering}}  \bibinfo{volume}{82} (\bibinfo{year}{2014}), \bibinfo{pages}{253--265}.
\newblock


\bibitem[Dassi et~al\mbox{.}(2015)]%
        {DASSI2015}
\bibfield{author}{\bibinfo{person}{Franco Dassi}, \bibinfo{person}{Hang Si}, \bibinfo{person}{Simona Perotto}, {and} \bibinfo{person}{Timo Streckenbach}.} \bibinfo{year}{2015}\natexlab{}.
\newblock \showarticletitle{Anisotropic Finite Element Mesh Adaptation via Higher Dimensional Embedding}.
\newblock \bibinfo{journal}{\emph{Procedia Engineering}}  \bibinfo{volume}{124} (\bibinfo{year}{2015}), \bibinfo{pages}{265--277}.
\newblock


\bibitem[Defferrard et~al\mbox{.}(2016)]%
        {Defferrard2016}
\bibfield{author}{\bibinfo{person}{Micha\"el Defferrard}, \bibinfo{person}{Xavier Bresson}, {and} \bibinfo{person}{Pierre Vandergheynst}.} \bibinfo{year}{2016}\natexlab{}.
\newblock \showarticletitle{Convolutional Neural Networks on Graphs with Fast Localized Spectral Filtering}.
\newblock  (\bibinfo{year}{2016}).
\newblock


\bibitem[Dobrzynski and Frey(2008)]%
        {dobrzynski2008}
\bibfield{author}{\bibinfo{person}{C{\'e}cile Dobrzynski} {and} \bibinfo{person}{Pascal Frey}.} \bibinfo{year}{2008}\natexlab{}.
\newblock \showarticletitle{Anisotropic Delaunay Mesh Adaptation for Unsteady Simulations}.
\newblock In \bibinfo{booktitle}{\emph{Proceedings of the 17th International Meshing Roundtable}}. \bibinfo{pages}{177--194}.
\newblock


\bibitem[Du and Wang(2005)]%
        {Du2005}
\bibfield{author}{\bibinfo{person}{Qiang Du} {and} \bibinfo{person}{Desheng Wang}.} \bibinfo{year}{2005}\natexlab{}.
\newblock \showarticletitle{Anisotropic Centroidal Voronoi Tessellations and Their Applications}.
\newblock \bibinfo{journal}{\emph{SIAM Journal on Scientific Computing}} \bibinfo{volume}{26}, \bibinfo{number}{3} (\bibinfo{year}{2005}), \bibinfo{pages}{737--761}.
\newblock


\bibitem[Fey and Lenssen(2019)]%
        {Fey/Lenssen/2019}
\bibfield{author}{\bibinfo{person}{Matthias Fey} {and} \bibinfo{person}{Jan~E. Lenssen}.} \bibinfo{year}{2019}\natexlab{}.
\newblock \showarticletitle{Fast Graph Representation Learning with {PyTorch Geometric}}. In \bibinfo{booktitle}{\emph{ICLR Workshop on Representation Learning on Graphs and Manifolds}}.
\newblock


\bibitem[Frey and Borouchaki(1999)]%
        {frey1999surface}
\bibfield{author}{\bibinfo{person}{Pascal~J Frey} {and} \bibinfo{person}{Houman Borouchaki}.} \bibinfo{year}{1999}\natexlab{}.
\newblock \showarticletitle{Surface Mesh Quality Evaluation}.
\newblock \bibinfo{journal}{\emph{International journal for numerical methods in engineering}} \bibinfo{volume}{45}, \bibinfo{number}{1} (\bibinfo{year}{1999}), \bibinfo{pages}{101--118}.
\newblock


\bibitem[Fu et~al\mbox{.}(2014)]%
        {fu2014anisotropic}
\bibfield{author}{\bibinfo{person}{Xiao-Ming Fu}, \bibinfo{person}{Yang Liu}, \bibinfo{person}{John Snyder}, {and} \bibinfo{person}{Baining Guo}.} \bibinfo{year}{2014}\natexlab{}.
\newblock \showarticletitle{Anisotropic Simplicial Meshing Using Local Convex Functions}.
\newblock \bibinfo{journal}{\emph{ACM Transactions on Graphics}} \bibinfo{volume}{33}, \bibinfo{number}{6} (\bibinfo{year}{2014}), \bibinfo{pages}{1--11}.
\newblock


\bibitem[Gao and Ji(2019)]%
        {gao2019graph}
\bibfield{author}{\bibinfo{person}{Hongyang Gao} {and} \bibinfo{person}{Shuiwang Ji}.} \bibinfo{year}{2019}\natexlab{}.
\newblock \showarticletitle{Graph u-nets}. In \bibinfo{booktitle}{\emph{international conference on machine learning}}. PMLR, \bibinfo{pages}{2083--2092}.
\newblock


\bibitem[Garland and Heckbert(1997)]%
        {garland1997surface}
\bibfield{author}{\bibinfo{person}{Michael Garland} {and} \bibinfo{person}{Paul~S Heckbert}.} \bibinfo{year}{1997}\natexlab{}.
\newblock \showarticletitle{Surface Simplification Using Quadric Error Metrics}. In \bibinfo{booktitle}{\emph{Proceedings of the Annual Conference on Computer Graphics and Interactive Techniques}}. \bibinfo{pages}{209--216}.
\newblock


\bibitem[Guennebaud et~al\mbox{.}(2010)]%
        {eigenweb}
\bibfield{author}{\bibinfo{person}{Ga\"{e}l Guennebaud}, \bibinfo{person}{Beno\^{i}t Jacob}, {et~al\mbox{.}}} \bibinfo{year}{2010}\natexlab{}.
\newblock \bibinfo{title}{Eigen v3}.
\newblock \bibinfo{howpublished}{http://eigen.tuxfamily.org}.
\newblock


\bibitem[Hamilton et~al\mbox{.}(2018)]%
        {hamilton2018inductive}
\bibfield{author}{\bibinfo{person}{William~L. Hamilton}, \bibinfo{person}{Rex Ying}, {and} \bibinfo{person}{Jure Leskovec}.} \bibinfo{year}{2018}\natexlab{}.
\newblock \bibinfo{title}{Inductive Representation Learning on Large Graphs}.
\newblock
\newblock


\bibitem[Hanocka et~al\mbox{.}(2019)]%
        {hanocka2019}
\bibfield{author}{\bibinfo{person}{Rana Hanocka}, \bibinfo{person}{Amir Hertz}, \bibinfo{person}{Noa Fish}, \bibinfo{person}{Raja Giryes}, \bibinfo{person}{Shachar Fleishman}, {and} \bibinfo{person}{Daniel Cohen-Or}.} \bibinfo{year}{2019}\natexlab{}.
\newblock \showarticletitle{MeshCNN: A Network with an Edge}.
\newblock \bibinfo{journal}{\emph{ACM Transactions on Graphics}} \bibinfo{volume}{38}, \bibinfo{number}{4} (\bibinfo{year}{2019}), \bibinfo{pages}{90:1--90:12}.
\newblock


\bibitem[Heckbert and Garland(1999)]%
        {Heckbert:1999}
\bibfield{author}{\bibinfo{person}{Paul~S Heckbert} {and} \bibinfo{person}{Michael Garland}.} \bibinfo{year}{1999}\natexlab{}.
\newblock \showarticletitle{Optimal Triangulation and Quadric-Based Surface Simplification}.
\newblock \bibinfo{journal}{\emph{Computational Geometry}} \bibinfo{volume}{14}, \bibinfo{number}{1--3} (\bibinfo{year}{1999}), \bibinfo{pages}{49--65}.
\newblock


\bibitem[Hu et~al\mbox{.}(2022)]%
        {Hu2021}
\bibfield{author}{\bibinfo{person}{Shi{-}Min Hu}, \bibinfo{person}{Zheng{-}Ning Liu}, \bibinfo{person}{Meng{-}Hao Guo}, \bibinfo{person}{Junxiong Cai}, \bibinfo{person}{Jiahui Huang}, \bibinfo{person}{Tai{-}Jiang Mu}, {and} \bibinfo{person}{Ralph~R. Martin}.} \bibinfo{year}{2022}\natexlab{}.
\newblock \showarticletitle{Subdivision-based Mesh Convolution Networks}.
\newblock \bibinfo{journal}{\emph{ACM Transactions on Graphics}} (\bibinfo{year}{2022}).
\newblock


\bibitem[Hu et~al\mbox{.}(2018)]%
        {Hu:2018}
\bibfield{author}{\bibinfo{person}{Yixin Hu}, \bibinfo{person}{Qingnan Zhou}, \bibinfo{person}{Xifeng Gao}, \bibinfo{person}{Alec Jacobson}, \bibinfo{person}{Denis Zorin}, {and} \bibinfo{person}{Daniele Panozzo}.} \bibinfo{year}{2018}\natexlab{}.
\newblock \showarticletitle{Tetrahedral Meshing in the Wild}.
\newblock \bibinfo{journal}{\emph{ACM Transactions on Graphics}} \bibinfo{volume}{37}, \bibinfo{number}{4} (\bibinfo{year}{2018}), \bibinfo{pages}{60:1--60:14}.
\newblock


\bibitem[Ioffe and Szegedy(2015)]%
        {ioffe2015batch}
\bibfield{author}{\bibinfo{person}{Sergey Ioffe} {and} \bibinfo{person}{Christian Szegedy}.} \bibinfo{year}{2015}\natexlab{}.
\newblock \showarticletitle{Batch normalization: Accelerating deep network training by reducing internal covariate shift}. In \bibinfo{booktitle}{\emph{International conference on machine learning}}. \bibinfo{pages}{448--456}.
\newblock


\bibitem[Ivrissimtzis et~al\mbox{.}(2004)]%
        {ivrissimtzis2004neural}
\bibfield{author}{\bibinfo{person}{Ioannis Ivrissimtzis}, \bibinfo{person}{Won-Ki Jeong}, \bibinfo{person}{Seungyong Lee}, \bibinfo{person}{Yunjin Lee}, {and} \bibinfo{person}{Hans-Peter Seidel}.} \bibinfo{year}{2004}\natexlab{}.
\newblock \showarticletitle{Neural Meshes: Surface Reconstruction with a Learning Algorithm}.
\newblock  (\bibinfo{year}{2004}).
\newblock


\bibitem[Jacobson et~al\mbox{.}(2018)]%
        {libigl}
\bibfield{author}{\bibinfo{person}{Alec Jacobson}, \bibinfo{person}{Daniele Panozzo}, {et~al\mbox{.}}} \bibinfo{year}{2018}\natexlab{}.
\newblock \bibinfo{title}{{libigl}: A simple {C++} geometry processing library}.
\newblock
\newblock
\newblock
\shownote{https://libigl.github.io/}.


\bibitem[Kipf and Welling(2017)]%
        {kipf2017semi}
\bibfield{author}{\bibinfo{person}{Thomas~N. Kipf} {and} \bibinfo{person}{Max Welling}.} \bibinfo{year}{2017}\natexlab{}.
\newblock \bibinfo{title}{Semi-Supervised Classification with Graph Convolutional Networks}.
\newblock
\newblock


\bibitem[Kostrikov et~al\mbox{.}(2018)]%
        {Kostrikov2018}
\bibfield{author}{\bibinfo{person}{Ilya Kostrikov}, \bibinfo{person}{Zhongshi Jiang}, \bibinfo{person}{Daniele Panozzo}, \bibinfo{person}{Denis Zorin}, {and} \bibinfo{person}{Burna Joan}.} \bibinfo{year}{2018}\natexlab{}.
\newblock \showarticletitle{Surface Networks}.
\newblock  (\bibinfo{year}{2018}).
\newblock


\bibitem[Kuiper(1955)]%
        {Kuiper:1955}
\bibfield{author}{\bibinfo{person}{Nicolaas~H Kuiper}.} \bibinfo{year}{1955}\natexlab{}.
\newblock \showarticletitle{On ${C}^1$-isometric embeddings {I}}. In \bibinfo{booktitle}{\emph{Proc. {N}ederl. {A}kad. {W}etensch. {S}er. {A}}}. \bibinfo{pages}{545--556}.
\newblock


\bibitem[L\'{e}vy(2015)]%
        {levy2015geogram}
\bibfield{author}{\bibinfo{person}{Bruno L\'{e}vy}.} \bibinfo{year}{2015}\natexlab{}.
\newblock \showarticletitle{Geogram}.
\newblock \bibinfo{journal}{\emph{GitHub Repository. URL: https://github. com/BrunoLevy/geogram}} (\bibinfo{year}{2015}).
\newblock


\bibitem[L{\'e}vy(2016)]%
        {levy_PCK}
\bibfield{author}{\bibinfo{person}{Bruno L{\'e}vy}.} \bibinfo{year}{2016}\natexlab{}.
\newblock \showarticletitle{{Robustness and Efficiency of Geometric Programs The Predicate Construction Kit (PCK)}}.
\newblock \bibinfo{journal}{\emph{Computer-Aided Design}}  \bibinfo{volume}{72} (\bibinfo{year}{2016}), \bibinfo{pages}{3--12}.
\newblock


\bibitem[L{\'e}vy and Bonneel(2013)]%
        {levy2013variational}
\bibfield{author}{\bibinfo{person}{Bruno L{\'e}vy} {and} \bibinfo{person}{Nicolas Bonneel}.} \bibinfo{year}{2013}\natexlab{}.
\newblock \showarticletitle{Variational Anisotropic Surface Meshing with Voronoi Parallel Linear Enumeration}.
\newblock In \bibinfo{booktitle}{\emph{Proceedings of the 21st International Meshing Roundtable}}. \bibinfo{pages}{349--366}.
\newblock


\bibitem[L\'{e}vy and Liu(2010)]%
        {levy2010LpCVT}
\bibfield{author}{\bibinfo{person}{Bruno L\'{e}vy} {and} \bibinfo{person}{Yang Liu}.} \bibinfo{year}{2010}\natexlab{}.
\newblock \showarticletitle{Lp Centroidal Voronoi Tessellation and Its Applications}.
\newblock \bibinfo{journal}{\emph{ACM Transactions on Graphics}} \bibinfo{volume}{29}, \bibinfo{number}{4} (\bibinfo{year}{2010}).
\newblock


\bibitem[Liu and Nocedal(1989)]%
        {liu1989limited}
\bibfield{author}{\bibinfo{person}{Dong~C Liu} {and} \bibinfo{person}{Jorge Nocedal}.} \bibinfo{year}{1989}\natexlab{}.
\newblock \showarticletitle{On the limited memory BFGS method for large scale optimization}.
\newblock \bibinfo{journal}{\emph{Mathematical programming}} \bibinfo{volume}{45}, \bibinfo{number}{1} (\bibinfo{year}{1989}), \bibinfo{pages}{503--528}.
\newblock


\bibitem[Lloyd(1982)]%
        {Lloyd1982}
\bibfield{author}{\bibinfo{person}{Stuart Lloyd}.} \bibinfo{year}{1982}\natexlab{}.
\newblock \showarticletitle{Least Squares Quantization in PCM}.
\newblock \bibinfo{journal}{\emph{IEEE Transactions on Information Theory}} \bibinfo{volume}{28}, \bibinfo{number}{2} (\bibinfo{year}{1982}), \bibinfo{pages}{129--137}.
\newblock


\bibitem[Loshchilov and Hutter(2019)]%
        {loshchilov2018decoupled}
\bibfield{author}{\bibinfo{person}{Ilya Loshchilov} {and} \bibinfo{person}{Frank Hutter}.} \bibinfo{year}{2019}\natexlab{}.
\newblock \showarticletitle{Decoupled Weight Decay Regularization}. In \bibinfo{booktitle}{\emph{International Conference on Learning Representations}}.
\newblock


\bibitem[Maas et~al\mbox{.}(2013)]%
        {maas2013rectifier}
\bibfield{author}{\bibinfo{person}{Andrew~L Maas}, \bibinfo{person}{Awni~Y Hannun}, \bibinfo{person}{Andrew~Y Ng}, {et~al\mbox{.}}} \bibinfo{year}{2013}\natexlab{}.
\newblock \showarticletitle{Rectifier Nonlinearities Improve Neural Network Acoustic Models}. In \bibinfo{booktitle}{\emph{Proc. icml}}, Vol.~\bibinfo{volume}{30}. \bibinfo{pages}{3}.
\newblock


\bibitem[Maruani et~al\mbox{.}(2024)]%
        {maruani2024ponq}
\bibfield{author}{\bibinfo{person}{Nissim Maruani}, \bibinfo{person}{Maks Ovsjanikov}, \bibinfo{person}{Pierre Alliez}, {and} \bibinfo{person}{Mathieu Desbrun}.} \bibinfo{year}{2024}\natexlab{}.
\newblock \showarticletitle{Po{NQ}: a Neural {QEM}-based Mesh Representation}. In \bibinfo{booktitle}{\emph{Proceedings of the IEEE/CVF Conference on Computer Vision and Pattern Recognition}}. \bibinfo{pages}{3647--3657}.
\newblock


\bibitem[Mescheder et~al\mbox{.}(2019)]%
        {mescheder2019occupancy}
\bibfield{author}{\bibinfo{person}{Lars Mescheder}, \bibinfo{person}{Michael Oechsle}, \bibinfo{person}{Michael Niemeyer}, \bibinfo{person}{Sebastian Nowozin}, {and} \bibinfo{person}{Andreas Geiger}.} \bibinfo{year}{2019}\natexlab{}.
\newblock \showarticletitle{Occupancy Networks: Learning {3D} Reconstruction in Function Space}. In \bibinfo{booktitle}{\emph{Proceedings of the IEEE/CVF conference on computer vision and pattern recognition}}. \bibinfo{pages}{4460--4470}.
\newblock


\bibitem[Narain et~al\mbox{.}(2012)]%
        {Narain:TOG2012}
\bibfield{author}{\bibinfo{person}{Rahul Narain}, \bibinfo{person}{Armin Samii}, {and} \bibinfo{person}{James~F O'brien}.} \bibinfo{year}{2012}\natexlab{}.
\newblock \showarticletitle{Adaptive Anisotropic Remeshing for Cloth Simulation}.
\newblock \bibinfo{journal}{\emph{ACM Transactions on Graphics}} \bibinfo{volume}{31}, \bibinfo{number}{6} (\bibinfo{year}{2012}), \bibinfo{pages}{147:1--147:10}.
\newblock


\bibitem[Nash(1954)]%
        {Nash:1954}
\bibfield{author}{\bibinfo{person}{John Nash}.} \bibinfo{year}{1954}\natexlab{}.
\newblock \showarticletitle{${C}^1$-Isometric Embeddings}.
\newblock \bibinfo{journal}{\emph{Annals of Mathematics}} \bibinfo{volume}{60}, \bibinfo{number}{3} (\bibinfo{year}{1954}), \bibinfo{pages}{383--396}.
\newblock


\bibitem[Nash(1956)]%
        {nash1956imbedding}
\bibfield{author}{\bibinfo{person}{John Nash}.} \bibinfo{year}{1956}\natexlab{}.
\newblock \showarticletitle{The Imbedding Problem for {R}iemannian Manifolds}.
\newblock \bibinfo{journal}{\emph{Annals of mathematics}} \bibinfo{volume}{63}, \bibinfo{number}{1} (\bibinfo{year}{1956}), \bibinfo{pages}{20--63}.
\newblock


\bibitem[Pang et~al\mbox{.}(2023)]%
        {Pang2023}
\bibfield{author}{\bibinfo{person}{Bo Pang}, \bibinfo{person}{Zhongtian Zheng}, \bibinfo{person}{Guoping Wang}, {and} \bibinfo{person}{Peng-Shuai Wang}.} \bibinfo{year}{2023}\natexlab{}.
\newblock \showarticletitle{Learning the Geodesic Embedding with Graph Neural Networks}.
\newblock \bibinfo{journal}{\emph{ACM Transactions on Graphics}} \bibinfo{volume}{42}, \bibinfo{number}{6} (\bibinfo{date}{Dec.} \bibinfo{year}{2023}), \bibinfo{pages}{1–12}.
\newblock


\bibitem[Panozzo et~al\mbox{.}(2014)]%
        {panozzo2014frame}
\bibfield{author}{\bibinfo{person}{Daniele Panozzo}, \bibinfo{person}{Enrico Puppo}, \bibinfo{person}{Marco Tarini}, {and} \bibinfo{person}{Olga Sorkine-Hornung}.} \bibinfo{year}{2014}\natexlab{}.
\newblock \showarticletitle{{Frame Fields:} Anisotropic and Non-Orthogonal Cross Fields}.
\newblock \bibinfo{journal}{\emph{ACM Transactions on Graphics}} \bibinfo{volume}{33}, \bibinfo{number}{4} (\bibinfo{year}{2014}), \bibinfo{pages}{134:1--134:11}.
\newblock


\bibitem[Pfaff et~al\mbox{.}(2020)]%
        {pfaff2020}
\bibfield{author}{\bibinfo{person}{Tobias Pfaff}, \bibinfo{person}{Meire Fortunato}, \bibinfo{person}{Alvaro Sanchez{-}Gonzalez}, {and} \bibinfo{person}{Peter~W. Battaglia}.} \bibinfo{year}{2020}\natexlab{}.
\newblock \showarticletitle{Learning Mesh-Based Simulation with Graph Networks}.
\newblock \bibinfo{journal}{\emph{CoRR}}  \bibinfo{volume}{abs/2010.03409} (\bibinfo{year}{2020}).
\newblock


\bibitem[Potamias et~al\mbox{.}(2022)]%
        {Potamias2022}
\bibfield{author}{\bibinfo{person}{Rolandos~Alexandros Potamias}, \bibinfo{person}{Stylianos Ploumpis}, {and} \bibinfo{person}{Stefanos Zafeiriou}.} \bibinfo{year}{2022}\natexlab{}.
\newblock \showarticletitle{Neural Mesh Simplification}.
\newblock  (\bibinfo{year}{2022}), \bibinfo{pages}{18562--18571}.
\newblock


\bibitem[Qi et~al\mbox{.}(2017a)]%
        {qi2017pointnet}
\bibfield{author}{\bibinfo{person}{Charles~R Qi}, \bibinfo{person}{Hao Su}, \bibinfo{person}{Kaichun Mo}, {and} \bibinfo{person}{Leonidas~J Guibas}.} \bibinfo{year}{2017}\natexlab{a}.
\newblock \showarticletitle{Point{N}et: Deep Learning on Point Sets for {3D} Classification and Segmentation}. In \bibinfo{booktitle}{\emph{Proceedings of the IEEE conference on computer vision and pattern recognition}}. \bibinfo{pages}{652--660}.
\newblock


\bibitem[Qi et~al\mbox{.}(2017b)]%
        {Qi_pointnet++}
\bibfield{author}{\bibinfo{person}{Charles~Ruizhongtai Qi}, \bibinfo{person}{Li Yi}, \bibinfo{person}{Hao Su}, {and} \bibinfo{person}{Leonidas~J Guibas}.} \bibinfo{year}{2017}\natexlab{b}.
\newblock \showarticletitle{PointNet++: Deep Hierarchical Feature Learning on Point Sets in a Metric Space}. In \bibinfo{booktitle}{\emph{Advances in Neural Information Processing Systems}}, \bibfield{editor}{\bibinfo{person}{I.~Guyon}, \bibinfo{person}{U.~Von Luxburg}, \bibinfo{person}{S.~Bengio}, \bibinfo{person}{H.~Wallach}, \bibinfo{person}{R.~Fergus}, \bibinfo{person}{S.~Vishwanathan}, {and} \bibinfo{person}{R.~Garnett}} (Eds.), Vol.~\bibinfo{volume}{30}. \bibinfo{publisher}{Curran Associates, Inc.}
\newblock


\bibitem[Rouxel-Labb{\'e} et~al\mbox{.}(2016)]%
        {rouxel2016discretized}
\bibfield{author}{\bibinfo{person}{Mael Rouxel-Labb{\'e}}, \bibinfo{person}{Mathijs Wintraecken}, {and} \bibinfo{person}{J-D Boissonnat}.} \bibinfo{year}{2016}\natexlab{}.
\newblock \showarticletitle{Discretized {R}iemannian Delaunay Triangulations}.
\newblock \bibinfo{journal}{\emph{Procedia engineering}}  \bibinfo{volume}{163} (\bibinfo{year}{2016}), \bibinfo{pages}{97--109}.
\newblock


\bibitem[Shapiro et~al\mbox{.}(1996)]%
        {shapiro1996}
\bibfield{author}{\bibinfo{person}{Paul~R Shapiro}, \bibinfo{person}{Hugo Martel}, \bibinfo{person}{Jens~V Villumsen}, {and} \bibinfo{person}{J~Michael Owen}.} \bibinfo{year}{1996}\natexlab{}.
\newblock \showarticletitle{Adaptive Smoothed Particle Hydrodynamics, with Application to Cosmology: Methodology}.
\newblock \bibinfo{journal}{\emph{Astrophysical Journal Supplement v. 103, p. 269}}  \bibinfo{volume}{103} (\bibinfo{year}{1996}), \bibinfo{pages}{269}.
\newblock


\bibitem[Sharp et~al\mbox{.}(2022)]%
        {Sharp2022DiffusionNet}
\bibfield{author}{\bibinfo{person}{Nicholas Sharp}, \bibinfo{person}{Souhaib Attaiki}, \bibinfo{person}{Keenan Crane}, {and} \bibinfo{person}{Maks Ovsjanikov}.} \bibinfo{year}{2022}\natexlab{}.
\newblock \showarticletitle{DiffusionNet: Discretization Agnostic Learning on Surfaces}.
\newblock \bibinfo{journal}{\emph{ACM Transactions on Graphics}} (\bibinfo{year}{2022}).
\newblock


\bibitem[Shimada et~al\mbox{.}(1997)]%
        {shimada1997}
\bibfield{author}{\bibinfo{person}{Kenji Shimada}, \bibinfo{person}{Atsushi Yamada}, \bibinfo{person}{Takayuki Itoh}, {et~al\mbox{.}}} \bibinfo{year}{1997}\natexlab{}.
\newblock \showarticletitle{Anisotropic Triangular Meshing of Parametric Surfaces via Close Packing of Ellipsoidal Bubbles}. In \bibinfo{booktitle}{\emph{6th International Meshing Roundtable}}, Vol.~\bibinfo{volume}{375}. \bibinfo{pages}{390}.
\newblock


\bibitem[Simpson(1994)]%
        {Simpson:1994}
\bibfield{author}{\bibinfo{person}{R~Bruce Simpson}.} \bibinfo{year}{1994}\natexlab{}.
\newblock \showarticletitle{Anisotropic Mesh Transformations and Optimal Error Control}.
\newblock \bibinfo{journal}{\emph{Applied Numerical Mathematics}} \bibinfo{volume}{14}, \bibinfo{number}{1--3} (\bibinfo{year}{1994}), \bibinfo{pages}{183--198}.
\newblock


\bibitem[Smirnov and Solomon(2021)]%
        {Smirnov2021}
\bibfield{author}{\bibinfo{person}{Dmitriy Smirnov} {and} \bibinfo{person}{Justin Solomon}.} \bibinfo{year}{2021}\natexlab{}.
\newblock \showarticletitle{HodgeNet: Learning Spectral Geometry on Triangle Meshes}.
\newblock \bibinfo{journal}{\emph{ACM Transactions on Graphics}}, Article \bibinfo{articleno}{166} (\bibinfo{date}{jul} \bibinfo{year}{2021}), \bibinfo{numpages}{11}~pages.
\newblock


\bibitem[Sumner and Popovi{\'c}(2004)]%
        {sumner2004deformation}
\bibfield{author}{\bibinfo{person}{Robert~W Sumner} {and} \bibinfo{person}{Jovan Popovi{\'c}}.} \bibinfo{year}{2004}\natexlab{}.
\newblock \showarticletitle{Deformation Transfer for Triangle Meshes}.
\newblock \bibinfo{journal}{\emph{ACM Transactions on Graphics}} \bibinfo{volume}{23}, \bibinfo{number}{3} (\bibinfo{year}{2004}), \bibinfo{pages}{399--405}.
\newblock


\bibitem[Sutherland and Hodgman(1974)]%
        {reentrant1974}
\bibfield{author}{\bibinfo{person}{Ivan~E. Sutherland} {and} \bibinfo{person}{Gary~W. Hodgman}.} \bibinfo{year}{1974}\natexlab{}.
\newblock \showarticletitle{Reentrant polygon clipping}.
\newblock \bibinfo{journal}{\emph{Commun. ACM}} \bibinfo{volume}{17}, \bibinfo{number}{1} (\bibinfo{date}{jan} \bibinfo{year}{1974}), \bibinfo{pages}{32–42}.
\newblock
\showISSN{0001-0782}
\urldef\tempurl%
\url{https://doi.org/10.1145/360767.360802}
\showDOI{\tempurl}


\bibitem[Valette et~al\mbox{.}(2008)]%
        {valette2008}
\bibfield{author}{\bibinfo{person}{S{\'e}bastien Valette}, \bibinfo{person}{Jean~Marc Chassery}, {and} \bibinfo{person}{R{\'e}my Prost}.} \bibinfo{year}{2008}\natexlab{}.
\newblock \showarticletitle{Generic Remeshing of 3D Triangular Meshes with Metric-Dependent Discrete Voronoi Diagrams}.
\newblock \bibinfo{journal}{\emph{IEEE Transactions on Visualization and Computer Graphics}} \bibinfo{volume}{14}, \bibinfo{number}{2} (\bibinfo{year}{2008}), \bibinfo{pages}{369--381}.
\newblock


\bibitem[Wang et~al\mbox{.}(2018b)]%
        {Wang2020}
\bibfield{author}{\bibinfo{person}{Nanyang Wang}, \bibinfo{person}{Yinda Zhang}, \bibinfo{person}{Zhuwen Li}, \bibinfo{person}{Yanwei Fu}, \bibinfo{person}{Wei Liu}, {and} \bibinfo{person}{Yu{-}Gang Jiang}.} \bibinfo{year}{2018}\natexlab{b}.
\newblock \showarticletitle{Pixel2{M}esh: Generating {3D} Mesh Models from Single {RGB} Images}.
\newblock \bibinfo{journal}{\emph{CoRR}}  \bibinfo{volume}{abs/1804.01654} (\bibinfo{year}{2018}).
\newblock


\bibitem[Wang et~al\mbox{.}(2017)]%
        {wang2017cnn}
\bibfield{author}{\bibinfo{person}{Peng-Shuai Wang}, \bibinfo{person}{Yang Liu}, \bibinfo{person}{Yu-Xiao Guo}, \bibinfo{person}{Chun-Yu Sun}, {and} \bibinfo{person}{Xin Tong}.} \bibinfo{year}{2017}\natexlab{}.
\newblock \showarticletitle{{O-CNN}: Octree-based convolutional neural networks for {3D} shape analysis}.
\newblock \bibinfo{journal}{\emph{ACM Transactions On Graphics}} \bibinfo{volume}{36}, \bibinfo{number}{4} (\bibinfo{year}{2017}), \bibinfo{pages}{1--11}.
\newblock


\bibitem[Wang et~al\mbox{.}(2018a)]%
        {wang2018adaptive}
\bibfield{author}{\bibinfo{person}{Peng-Shuai Wang}, \bibinfo{person}{Chun-Yu Sun}, \bibinfo{person}{Yang Liu}, {and} \bibinfo{person}{Xin Tong}.} \bibinfo{year}{2018}\natexlab{a}.
\newblock \showarticletitle{Adaptive {O-CNN}: A patch-based deep representation of {3D} shapes}.
\newblock \bibinfo{journal}{\emph{ACM Transactions on Graphics}} \bibinfo{volume}{37}, \bibinfo{number}{6} (\bibinfo{year}{2018}), \bibinfo{pages}{1--11}.
\newblock


\bibitem[Xiao et~al\mbox{.}(2020)]%
        {xiao2020survey}
\bibfield{author}{\bibinfo{person}{Yun-Peng Xiao}, \bibinfo{person}{Yu-Kun Lai}, \bibinfo{person}{Fang-Lue Zhang}, \bibinfo{person}{Chunpeng Li}, {and} \bibinfo{person}{Lin Gao}.} \bibinfo{year}{2020}\natexlab{}.
\newblock \showarticletitle{A Survey on Deep Geometry Learning: from a Representation Perspective}.
\newblock \bibinfo{journal}{\emph{Computational Visual Media}}  \bibinfo{volume}{6} (\bibinfo{year}{2020}), \bibinfo{pages}{113--133}.
\newblock


\bibitem[Xu et~al\mbox{.}(2024)]%
        {xu2024cwf}
\bibfield{author}{\bibinfo{person}{Rui Xu}, \bibinfo{person}{Longdu Liu}, \bibinfo{person}{Ningna Wang}, \bibinfo{person}{Shuangmin Chen}, \bibinfo{person}{Shiqing Xin}, \bibinfo{person}{Xiaohu Guo}, \bibinfo{person}{Zichun Zhong}, \bibinfo{person}{Taku Komura}, \bibinfo{person}{Wenping Wang}, {and} \bibinfo{person}{Changhe Tu}.} \bibinfo{year}{2024}\natexlab{}.
\newblock \showarticletitle{{CWF}: Consolidating Weak Features in High-quality Mesh Simplification}.
\newblock \bibinfo{journal}{\emph{ACM Transactions on Graphics}} \bibinfo{volume}{43}, \bibinfo{number}{4} (\bibinfo{year}{2024}), \bibinfo{pages}{1--14}.
\newblock


\bibitem[Yan et~al\mbox{.}(2009)]%
        {yan2009RVD}
\bibfield{author}{\bibinfo{person}{Dong-Ming Yan}, \bibinfo{person}{Bruno L\'{e}vy}, \bibinfo{person}{Yang Liu}, \bibinfo{person}{Feng Sun}, {and} \bibinfo{person}{Wenping Wang}.} \bibinfo{year}{2009}\natexlab{}.
\newblock \showarticletitle{Isotropic Remeshing with Fast and Exact Computation of Restricted Voronoi Diagram}.
\newblock \bibinfo{journal}{\emph{Computer Graphics Forum}} \bibinfo{volume}{28}, \bibinfo{number}{5} (\bibinfo{year}{2009}), \bibinfo{pages}{1445--1454}.
\newblock


\bibitem[Zhong et~al\mbox{.}(2013)]%
        {Particle2013}
\bibfield{author}{\bibinfo{person}{Zichun Zhong}, \bibinfo{person}{Xiaohu Guo}, \bibinfo{person}{Wenping Wang}, \bibinfo{person}{Bruno L\'{e}vy}, \bibinfo{person}{Feng Sun}, \bibinfo{person}{Yang Liu}, {and} \bibinfo{person}{Weihua Mao}.} \bibinfo{year}{2013}\natexlab{}.
\newblock \showarticletitle{Particle-Based Anisotropic Surface Meshing}.
\newblock \bibinfo{journal}{\emph{ACM Transactions on Graphics}} \bibinfo{volume}{32}, \bibinfo{number}{4} (\bibinfo{year}{2013}).
\newblock


\bibitem[Zhong et~al\mbox{.}(2014)]%
        {zhong2014anisotropic}
\bibfield{author}{\bibinfo{person}{Zichun Zhong}, \bibinfo{person}{Liang Shuai}, \bibinfo{person}{Miao Jin}, {and} \bibinfo{person}{Xiaohu Guo}.} \bibinfo{year}{2014}\natexlab{}.
\newblock \showarticletitle{Anisotropic Surface Meshing with Conformal Embedding}.
\newblock \bibinfo{journal}{\emph{Graphical models}} \bibinfo{volume}{76}, \bibinfo{number}{5} (\bibinfo{year}{2014}), \bibinfo{pages}{468--483}.
\newblock


\bibitem[Zhong et~al\mbox{.}(2018)]%
        {Embedding2018}
\bibfield{author}{\bibinfo{person}{Zichun Zhong}, \bibinfo{person}{Wenping Wang}, \bibinfo{person}{Bruno L\'{e}vy}, \bibinfo{person}{Jing Hua}, {and} \bibinfo{person}{Xiaohu Guo}.} \bibinfo{year}{2018}\natexlab{}.
\newblock \showarticletitle{Computing a High-Dimensional {E}uclidean Embedding from an Arbitrary Smooth {R}iemannian Metric}.
\newblock \bibinfo{journal}{\emph{ACM Transactions on Graphics}} \bibinfo{volume}{37}, \bibinfo{number}{4} (\bibinfo{year}{2018}).
\newblock


\bibitem[Zhou et~al\mbox{.}(2016)]%
        {Zhou:2016:MASG}
\bibfield{author}{\bibinfo{person}{Qingnan Zhou}, \bibinfo{person}{Eitan Grinspun}, \bibinfo{person}{Denis Zorin}, {and} \bibinfo{person}{Alec Jacobson}.} \bibinfo{year}{2016}\natexlab{}.
\newblock \showarticletitle{Mesh Arrangements for Solid Geometry}.
\newblock \bibinfo{journal}{\emph{ACM Transactions on Graphics}} \bibinfo{volume}{35}, \bibinfo{number}{4} (\bibinfo{year}{2016}).
\newblock


\end{thebibliography}
